\documentclass[11pt]{article}

\usepackage[utf8]{inputenc}
\usepackage[T1]{fontenc}

\usepackage{arxiv}

\usepackage{amsmath}
\usepackage{amssymb}
\usepackage{graphicx}
\usepackage{booktabs}
\usepackage{listings}
\usepackage{subcaption}
\usepackage{enumitem}
\usepackage{natbib}

\usepackage{hyperref}
\usepackage{url}
\hypersetup{
  colorlinks=true,
  linkcolor=arxivaccent,
  citecolor=arxivaccent,
  urlcolor=arxivaccent,
  filecolor=arxivaccent,
  breaklinks=true,
}

\definecolor{codebg}{rgb}{0.95,0.95,0.97}
\definecolor{codegreen}{rgb}{0.0,0.5,0.0}
\definecolor{codegray}{rgb}{0.5,0.5,0.5}
\definecolor{codepurple}{rgb}{0.58,0,0.82}
\definecolor{codeblue}{rgb}{0.0,0.0,0.7}

\lstdefinestyle{pythonstyle}{
    backgroundcolor=\color{codebg},
    basicstyle=\ttfamily\small,
    breakatwhitespace=false,
    breaklines=true,
    captionpos=b,
    commentstyle=\color{codegreen},
    keywordstyle=\color{codeblue}\bfseries,
    numberstyle=\tiny\color{codegray},
    stringstyle=\color{codepurple},
    language=Python,
    showstringspaces=false,
    numbers=left,
    numbersep=5pt,
    frame=single,
    rulecolor=\color{codegray},
    tabsize=4,
    morekeywords={dataclass, List, Optional, date, assert, Dict, field}
}

\lstdefinestyle{treestyle}{
    basicstyle=\ttfamily\small,
    frame=none,
    numbers=none,
    breaklines=false,
    backgroundcolor=\color{codebg},
}

\definecolor{convbg}{rgb}{0.98,0.97,0.92}
\definecolor{speakerblue}{rgb}{0.10,0.30,0.60}
\lstdefinestyle{conversationstyle}{
    backgroundcolor=\color{convbg},
    basicstyle=\ttfamily\footnotesize,
    breakatwhitespace=false,
    breaklines=true,
    captionpos=b,
    language={},          
    keywordstyle={},
    stringstyle={},
    commentstyle={},
    showstringspaces=false,
    numbers=none,
    frame=single,
    rulecolor=\color{codegray},
    tabsize=2,
}

\definecolor{alertbg}{rgb}{1.00,0.96,0.94}
\definecolor{alertred}{rgb}{0.70,0.10,0.10}
\lstdefinestyle{alertstyle}{
    backgroundcolor=\color{alertbg},
    basicstyle=\ttfamily\footnotesize,
    breakatwhitespace=false,
    breaklines=true,
    captionpos=b,
    commentstyle=\color{codegreen},
    keywordstyle=\color{alertred}\bfseries,
    showstringspaces=false,
    numbers=none,
    frame=single,
    rulecolor=\color{alertred},
    tabsize=2,
    morekeywords={CRITICAL,WARNING,INFO,ALERT,DRUG-ALLERGY,CONFLICT},
}

\title{User as Code: Executable Memory for Personalized Agents}

\author{
  Bojie Li \\
  Pine AI
}

\date{}
\runningtitle{User as Code: Executable Memory for Personalized Agents}

\begin{document}
\maketitle

\begin{abstract}
A personalized AI agent needs a \emph{user memory}: a persistent model of who the user is, accumulated across many conversations and consulted on each new one. Today this memory is almost always stored as unstructured text, a knowledge graph, or a flat store of extracted facts, and consulted by \emph{retrieval}---fetching the stored entries most similar to the current request. Such ``bag-of-facts'' memory recalls individual facts well, but because storing a fact and acting on it are separate steps, it struggles to resolve contradictions, aggregate over many records, or enforce logical rules. We argue that user memory should instead be \emph{executable}. We introduce \textbf{User as Code} (UaC), a paradigm in which an agent's model of a user is a living software project: typed Python objects hold the user's state, and ordinary Python functions encode the rules that govern it, so that representing the user and reasoning about the user happen in one medium an interpreter can run. The enabling mechanism is a \emph{two-phase pipeline} that turns raw conversation into this code---an append-only log that never discards a fact, periodically checkpointed into structured, typed code---a design long used in database systems and, to our knowledge, applied here for the first time to LLM memory.

This shift changes what memory can do. On standard long-term conversation benchmarks, UaC is competitive with both a full-context upper bound and the strongest prior memory systems on ordinary factual recall (78.8\% on LOCOMO). Its advantage emerges where representation matters most. On aggregate questions over a user's history---\emph{``how many international trips did I take last year?''}---retrieval-based memory collapses (6--43\%) while UaC stays near-perfect (99\%), because the answer is a one-line computation over typed state rather than a search over text. And because its rules, once written, execute deterministically whenever the state changes, UaC can surface unsolicited, safety-critical alerts---such as a newly prescribed drug that conflicts with an allergy recorded months earlier---a proactive capability that query-driven memory cannot provide at all. The one-time cost of structuring the code is repaid after only a few queries against the same user.
\end{abstract}

\begin{center}
\small
Code: \url{https://github.com/19PINE-AI/user-as-code} \\[2pt]
Website: \url{https://01.me/research/user-as-code}
\end{center}
\vspace{-0.6em}

\section{Introduction}
\label{sec:intro}

Personalized AI agents need a \emph{user memory}: a model of the user that accumulates across sessions, not a transcript of every utterance. User memory tasks fall into three tiers: (1)~\textbf{Basic Recall}---retrieving an unambiguous fact (\emph{``what is my passport number?''}); (2)~\textbf{Multi-session Retrieval}---reasoning over information scattered across, and sometimes contradicted by, later sessions (\emph{``which doctor did I see for my allergies?''} when the relevant turns are months apart); and (3)~\textbf{Active Service}---producing anticipatory, unsolicited help, e.g.\ warning that a passport will expire within 180 days of a booked trip \emph{before the user thinks to ask}. Existing benchmarks~\citep{maharana2024locomo, wu2025longmemeval, hu2025memoryagentbench} test the first two but miss the \emph{initiation asymmetry} central to (3): the system must generate alerts \emph{without being asked}, triggered by state changes rather than queries.

Current memory systems---vector databases, JSON stores, knowledge graphs, flat fact extractors---treat memory as a loose collection of facts. This creates two problems. Conflicting facts (``I love cilantro'' vs.\ ``I actually hate cilantro'') coexist unresolved without version history. And even the most advanced formats cannot express executable rules like ``if passport expires within 180 days of travel date, raise an alert.''

Free-text and fact-store formats separate \emph{representation} from \emph{verification}. Any representation a Python interpreter can read directly---typed objects, JSON dicts, even strings with a parser---collapses the two: \texttt{passport.expiry\_date = date(2025, 2, 18)} can be stored \emph{and} compared in one medium (Listings~\ref{lst:travel-state}--\ref{lst:constraint}). The analytical benchmark confirms this: a baseline that loads all raw records and can execute Python over them (Full~Context~+~REPL) matches UaC on aggregate queries (Section~\ref{sec:analytical}), so ``code-as-representation'' is not by itself a novel claim. \emph{UaC's contribution is the two-phase pipeline that converts unstructured conversation into the code-readable representation in the first place}---without it, every aggregate query pays LLM parsing cost over noisy dialogue. The pipeline shape (append-only log + periodic checkpoint) borrows from database systems; the novelty is applying it to LLM memory with typed Python as the checkpoint medium.

\begin{figure}[htbp]
\begin{lstlisting}[caption={Three capabilities UaC provides, all against the same auto-generated typed state. (a)~Recall is attribute access---what every memory system gives you. (b)~Analytical aggregate is a one-line expression---where retrieval-based systems collapse (Section~\ref{sec:analytical}). (c)~Constraint is a boolean check the interpreter fires on every state change---the basis of the proactive alerts in Section~\ref{sec:active-service}.},label={lst:tiers}]
# Typed user state (excerpt, auto-generated by Phase 2):
passport = PassportInfo(number="AB1234567",
                        expiry_date=date(2025, 2, 18))
trips = [Trip(destination="Tokyo",       departure_date=date(2025, 1, 15),
              is_international=True),
         Trip(destination="Mexico City", departure_date=date(2025, 3, 10),
              is_international=True),
         Trip(destination="Portland",    departure_date=date(2025, 4, 22),
              is_international=False)]

# (a) Recall -- one attribute access:
>>> passport.number
'AB1234567'

# (b) Analytical aggregate -- one-line expression over the collection:
>>> sum(1 for t in trips if t.is_international
                          and t.departure_date.year == 2025)
2

# (c) Constraint -- boolean check; interpreter fires the alert:
>>> (passport.expiry_date - trips[0].departure_date).days >= 180
False                # 34 days; passport renewal needed before Tokyo
\end{lstlisting}
\end{figure}

\textbf{Three things memory-as-code unlocks.} Listing~\ref{lst:tiers} shows three capabilities, all against the same typed state, all one-line Python. These refine the task tiers above: (a) is Basic Recall, (c) is Active Service, and (b)---analytical inference---is a distinct capability beyond plain multi-session retrieval, isolated by its own benchmark (Section~\ref{sec:analytical}). (a)~\textbf{Recall}---answering \emph{``what is my passport number?''}---is attribute access. (b)~\textbf{Analytical inference}---aggregate questions over the entire history, such as \emph{``how many international trips did I take in 2025?''} or \emph{``what was my average dining spend by cuisine last year?''}---is a list comprehension: trivial in Python, but structurally lossy under top-$k$ retrieval, which only ever sees a handful of records at once (UaC 99\% vs.\ MemMachine 43\%, Section~\ref{sec:analytical}). (c)~\textbf{Active Service}---an unsolicited, state-triggered alert, such as flagging that a passport expires within 180 days of a booked trip, or that a newly prescribed antibiotic conflicts with a penicillin allergy logged months earlier---is a boolean check the interpreter runs deterministically at every state change, with no user query to trigger it. Listings~\ref{lst:teaser-conv}--\ref{lst:teaser-alert} expand (c) into a life-safety walkthrough where the relevant facts arrive ten months apart.

\textbf{A deeper example: cross-session Active Service.} The check in Listing~\ref{lst:tiers}(c) is small because the relevant facts sit in one state object. The harder case is when they arrive in different conversations and only meet inside the schema. Listings~\ref{lst:teaser-conv}--\ref{lst:teaser-alert} show this end-to-end. Two utterances ten months apart (Listing~\ref{lst:teaser-conv}) distill into an append-only fact list and a typed Python state (Listing~\ref{lst:teaser-state}); they never co-occur in any retrieval window, but \texttt{drug\_class="penicillin"} on both an \texttt{Allergy} and a \texttt{Medication} is a deterministic boolean check. The coding agent writes a 12-line constraint (Listing~\ref{lst:teaser-constraint}); the interpreter runs it; the alert (Listing~\ref{lst:teaser-alert}) appears in the next session's manifest before the user asks anything. Free-text retrieval would need to either know that amoxicillin is a penicillin (uncertain under retrieval pressure) or rank a year-old allergy turn highly against a sinus-infection query---neither is reliable.

\begin{figure}[htbp]
\begin{lstlisting}[style=conversationstyle,caption={Two conversation turns, ten months apart.},label={lst:teaser-conv}]
SESSION 12 -- DATE: 2024-03-01
Jessica: Saw Dr. Park about my seasonal allergies, she put me on
         cetirizine 10mg. Reminder -- I'm DEATHLY allergic to
         penicillin, anaphylaxis. Last time I almost died.

SESSION 47 -- DATE: 2025-01-10  (ten months later)
Jessica: Quick log: Dr. Robert Chen prescribed amoxicillin 500mg,
         three times a day for ten days, for the sinus infection.
\end{lstlisting}
\end{figure}

\begin{figure}[htbp]
\begin{lstlisting}[caption={Phase~1 facts and Phase~2 typed state. The shared \texttt{drug\_class="penicillin"} key anchors the cross-session link.},label={lst:teaser-state}]
# Phase 1 -- append-only fact list (never overwritten)
facts = [
    "[2024-03-01] User has SEVERE penicillin allergy (anaphylaxis)",
    "[2024-03-01] Dr. Park prescribed cetirizine 10mg daily",
    "[2025-01-10] Dr. Chen prescribed amoxicillin 500mg, 3x/day, "
    "10 days, sinus infection",
]

# Phase 2 -- structured Python (regenerated from full fact corpus)
medical_profile = MedicalProfile(
    allergies=[Allergy(allergen="Penicillin", severity="severe",
        reaction="Anaphylaxis", drug_class="penicillin")],
    current_medications=[
        Medication(name="Cetirizine", drug_class="antihistamine",
            dosage="10mg", prescriber="Dr. Park",
            start_date=date(2024, 3, 1)),
        Medication(name="Amoxicillin", drug_class="penicillin",
            dosage="500mg", prescriber="Dr. Chen",
            start_date=date(2025, 1, 10)),
    ],
)
\end{lstlisting}
\end{figure}

\begin{figure}[htbp]
\begin{lstlisting}[caption={The constraint the coding agent writes once, against the typed state. The interpreter executes it deterministically -- no LLM at check time.},label={lst:teaser-constraint}]
def check_drug_allergy(profile: MedicalProfile) -> list[Alert]:
    alerts = []
    for med in profile.current_medications:
        for allergy in profile.allergies:
            if med.drug_class == allergy.drug_class:
                alerts.append(Alert(
                    severity="critical", domain="health",
                    message=(f"DRUG-ALLERGY CONFLICT: {med.name} "
                            f"({med.drug_class}-class), prescribed "
                            f"{med.start_date} by {med.prescriber}; "
                            f"patient has {allergy.severity} "
                            f"{allergy.allergen} allergy "
                            f"({allergy.reaction}).")))
    return alerts
\end{lstlisting}
\end{figure}

\begin{figure}[htbp]
\begin{lstlisting}[style=alertstyle,caption={Alert surfaced in \texttt{ACTIVE\_ALERTS} at the start of every subsequent session. No retrieval, no question asked -- the constraint fired the moment the state was updated.},label={lst:teaser-alert}]
[CRITICAL/health]  DRUG-ALLERGY CONFLICT: Amoxicillin
(penicillin-class), prescribed 2025-01-10 by Dr. Chen;
patient has severe Penicillin allergy (Anaphylaxis).
\end{lstlisting}
\end{figure}

\begin{figure}[htbp]
\centering
\includegraphics[width=0.98\textwidth]{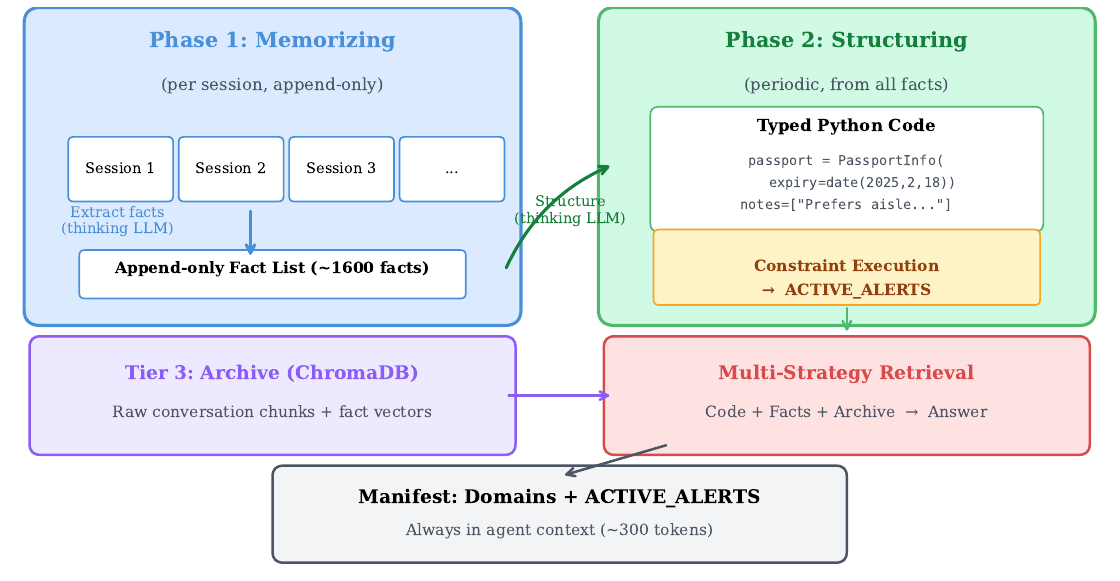}
\caption{User as Code two-phase architecture. Phase~1 extracts facts from each session (append-only, never overwritten). Phase~2 periodically structures all accumulated facts into typed Python dataclasses. Multi-strategy retrieval combines structured code, fact vectors, and raw archive. Constraint execution produces pre-computed alerts surfaced in the manifest.}
\label{fig:architecture}
\end{figure}

Concretely, we model user memory as a self-evolving software project---typed dataclasses, domain packages, executable constraints---driven by a two-phase architecture (Figure~\ref{fig:architecture}). Append-only fact extraction never loses information; periodic code structuring organizes facts into typed Python. Together they enable a generate--verify--review loop: the coding agent writes constraints, the interpreter verifies them deterministically, and results surface as pre-computed alerts in a compact manifest. Our contributions:
\begin{itemize}[leftmargin=*]
\item \textbf{A two-phase memory architecture} reaching 78.8\% on the full 10-conversation LOCOMO benchmark (600 QAs), within 1.0pp of the full-context upper bound (79.8\%, McNemar $p{=}0.65$), and significantly ahead of same-backbone SOTA reimplementations: MemMachine 72.7\%, Hindsight 69.7\%, EverMemOS 55.5\% (all $p<0.005$). The same architecture reaches 100\% proactive alert detection on 40 standard Active Service scenarios (85\% on 20 hard scenarios).
\item \textbf{An Active Service benchmark} testing proactive alerting across 60 scenarios in 5 constraint categories---the first benchmark we are aware of to measure memory-triggered alerts initiated without a user query.
\item \textbf{A new analytical-inference benchmark} (100 cases, 10 record types, $N{\in}\{20, 50, 100, 200, 500\}$) exposing a structural gap between code-executable representations (UaC 99\%, FC+REPL 100\%) and retrieval-based ones (MemMachine 43\%, Mem0 6\%). Token instrumentation shows UaC's structuring pays back after 3--11 repeated queries against the same user state (3 at the largest, $N{=}500$, state size) and amortizes to $\sim$15$\times$ cheaper than reloading raw records.
\item \textbf{Three orthogonal ablations} attributing the architecture's gains: (i)~append-only fact extraction is the single decisive recall mechanism (+19pp over a code-only baseline), and two-phase separation adds typed structure without the information loss of incremental code rewrites (+12.3pp over incremental code); (ii)~a leave-one-out channel study shows the typed state is roughly neutral for LOCOMO recall ($-1.3$pp, $p{=}0.67$) but decisive for the analytical and constraint tiers retrieval cannot serve; (iii)~Modularity/Progressive Disclosure reduces prompt-token cost 14.9$\times$ with no accuracy loss on 500-record states.
\item \textbf{Cross-LLM and cross-judge robustness checks.} Replacing Gemini~3 Flash with GPT-5.4 throughout the pipeline yields a statistical tie on a 120-QA LOCOMO subset ($p{=}0.82$). Rejudging all 7{,}700 predictions under Claude Opus~4.7 preserves the ranking with high inter-judge agreement (Cohen's $\kappa \geq 0.74$) on every system.
\item \textbf{Open-source implementation} with same-backbone comparisons against five memory systems (Mem0, A-MEM, MemMachine, EverMemOS, Hindsight) plus full-context baselines on LOCOMO, LongMemEval, Active Service, Analytical Inference, and Modularity.
\end{itemize}

\section{Related Work}
\label{sec:related}

LLM agent memory has grown rapidly in 2025--2026, with 20+ systems, 15+ benchmarks, and several surveys~\citep{hu2025memoryage, du2026memorysurvey, yang2026graphmemory, wu2025humantoai, zhang2024memorysurvey}. The broader goal these systems serve---adapting a general model to an individual---is the subject of a parallel and equally active literature on LLM personalization~\citep{zhang2024personalization, liu2025personalizedllm}, increasingly framed around personalized \emph{agents} whose profile, memory, planning, and actions are all tailored to one user~\citep{xu2026personalizedagents}. UaC sits at the intersection: it is a memory mechanism whose purpose is personalization. We organize related work around three gaps UaC addresses, after first positioning it against the line of work that treats code---rather than text---as the medium an LLM reads and writes.

\subsection{Code as a Representation for LLMs}
\label{sec:related-code}

A substantial line of work shows that generated code is a better substrate than free text when a task demands precise computation, composition, or verification. \textbf{Code as reasoning}: Program-of-Thoughts~\citep{chen2023pot} and PAL~\citep{gao2023pal} offload arithmetic and logic to a Python interpreter, decoupling \emph{what to compute} from \emph{computing it correctly}, and Chain-of-Code~\citep{li2024chainofcode} extends this to semantic sub-tasks by interleaving real execution with LM-emulated execution. \textbf{Code as structured representation}: even when the end task involves no source code, framing it as code generation makes LLMs better structured reasoners---CoCoGen~\citep{madaan2022cocogen} shows code-LMs emit cleaner structured (graph) outputs than NL-LMs, the same observation that motivates representing user state as typed Python rather than prose. \textbf{Code as action}: CodeAct~\citep{wang2024codeact} unifies an agent's action space into executable Python, outperforming JSON/text tool calls; this contrasts with the text-and-action interleaving of ReAct~\citep{yao2023react} and the natural-language self-reflection of Reflexion~\citep{shinn2023reflexion}. \textbf{Code as a skill/tool library}: Voyager~\citep{wang2024voyager} stores acquired behaviors as retrievable executable functions; LLMs-as-Tool-Makers~\citep{cai2023latm} and TroVE~\citep{wang2024trove} have models author and reuse their own verifiable Python utilities; and recent work distills agent experience into reusable procedural skills~\citep{mi2026skillpro}. \textbf{Code as policy}: Code-as-Policies~\citep{liang2023codeaspolicies} generates control programs for embodied agents, and neurosymbolic systems synthesize machine-checkable constraints from natural language for high-stakes domains~\citep{akinfaderin2025verafi}.

\textbf{How UaC differs.} These systems apply code to \emph{reasoning}, \emph{actions}, \emph{skills}, \emph{tools}, or \emph{policies}---and even when they target a structured \emph{representation}~\citep{madaan2022cocogen}, the artifact is produced to accomplish one task and then discarded. UaC applies the same insight to the one thing they all leave in text: the \emph{persistent state of the user}. The user's profile, history, and the rules that govern them become a versioned directory of typed Python that an interpreter computes over directly, rather than a transcript an LLM re-reads each turn. The benefits these works document for transient code---determinism, composition, verifiability---we claim for long-lived memory.

\subsection{Gap 1: Representation Separates Storage from Verification}

Existing memory systems store user state in formats that cannot be directly executed. Early agent-memory designs such as Generative Agents~\citep{park2023generative} introduced a memory stream with periodic reflection, and most current systems inherit this store-then-retrieve shape. \textbf{Flat fact systems} like Mem0~\citep{chhikara2025mem0} extract atomic facts with a CRUD lifecycle. \textbf{Structured-note systems} like A-MEM~\citep{xu2025amem} use Zettelkasten-inspired notes (atomic notes that cross-link to one another) with dynamic linking; ENGRAM~\citep{patel2025engram} shows that careful typing plus dense retrieval can match more complex architectures. \textbf{Knowledge-graph approaches} like Zep/Graphiti~\citep{rasmussen2025zep} build temporal knowledge graphs with bitemporal modeling (tracking both when a fact was true and when it was recorded); MAGMA~\citep{jiang2026magma} adds multi-graph traversal. \textbf{Hybrid architectures} like Hindsight~\citep{latimer2025hindsight} maintain four logical networks (91.4\% on LongMemEval); MemMachine~\citep{wang2026memmachine} stores entire episodes (91.7\% on LOCOMO). \textbf{OS-inspired systems} like MemGPT/Letta~\citep{packer2023memgpt}, MemOS~\citep{li2025memos}, and MemoryOS~\citep{kang2025memoryos} treat memory paging as an OS primitive: the agent edits text-based ``main context'' and ``recall storage'' pages through a custom API (Letta's \texttt{core\_memory\_replace}~/~\texttt{archival\_memory\_insert}), but the page contents remain unstructured text---the OS is around the memory, not in it. \textbf{Agent-editable instructions} are taken furthest by LangMem~\citep{langmem2025}, which lets the agent rewrite its own system-prompt directives over time; this is the closest existing system to ``executable memory,'' but the medium remains natural language interpreted by the LLM at read time, not typed objects an interpreter can run.

\textbf{How UaC differs.} UaC commits to typed Python for the \emph{state itself}, not just for instructions or paging machinery. Concretely: (i)~user state is a directory of Python dataclass instances, not text pages---an interpreter indexes, groups, and computes over it without reparsing; (ii)~persistent ``rules'' are Python functions over that state (e.g., \texttt{check\_drug\_allergy(profile) -> list[Alert]}), not natural-language directives the LLM re-interprets each call; (iii)~MemGPT-style paging and LangMem-style instruction editing remain available on top---each domain is a module, each constraint a callable---but neither is required for the headline capabilities. The closest ``state-as-code'' work outside the memory community is constraint-as-code in policy languages (e.g., Rego/OPA) and the LLM-generated, machine-checkable policies of neurosymbolic agents~\citep{liang2023codeaspolicies, akinfaderin2025verafi}; UaC differs in that schemas, instances, \emph{and} constraints are all LLM-generated over evolving user state, not human-authored and frozen, nor synthesized fresh per task and discarded.

\subsection{Gap 2: No Benchmark Tests Proactive Memory-Triggered Alerting}

LOCOMO~\citep{maharana2024locomo} provides 1,986 QA pairs; LongMemEval~\citep{wu2025longmemeval} tests 500 questions across six question types. LoCoMo-Plus~\citep{li2026locomoplus} tests implicit constraints; MemoryArena~\citep{he2026memoryarena} shows LOCOMO-saturated agents fail in agentic settings. Proactive assistance itself is an established goal---Proactive Agent~\citep{lu2025proactive} formalizes anticipating user needs without explicit requests---and ProAgentBench~\citep{tang2026proagentbench} tests proactive task assistance, but neither targets memory-triggered alerting. PersistBench~\citep{pulipaka2026persistbench} identifies memory safety risks. To our knowledge no existing benchmark tests the \emph{initiation asymmetry} central to Active Service---the system must produce alerts in the absence of a user query, triggered by state changes alone---which motivates the new Active Service benchmark in Section~\ref{sec:active-service}.

\subsection{Gap 3: Memory Operations Are Not Yet Optimized End-to-End}

Memory-R1~\citep{yan2025memoryr1} trains memory policies with GRPO (+48\% F1). AgeMem~\citep{yu2026agemem} learns proactive summarization through progressive RL. Mem-alpha~\citep{wang2025memalpha} demonstrates generalization to 400K+ tokens. Cognitively inspired systems include FadeMem~\citep{wei2026fademem} (biologically inspired forgetting), LightMem~\citep{fang2025lightmem} (Atkinson-Shiffrin model), and EverMemOS~\citep{hu2026evermemos} (engram-inspired lifecycles, 93.05\% on LOCOMO). All train over custom memory APIs; UaC's file-system-based architecture makes memory operations native file actions, which is directly compatible with the same RL machinery.

\section{Methodology: The User as Code Architecture}
\label{sec:method}

\textbf{Two pipelines.} The architecture has two halves, and the rest of this section is organized around them. \emph{Extraction} (Section~\ref{sec:extraction}) converts raw multi-session conversation into code, in two phases---Memorize, then Structure. \emph{Retrieval and use} (Section~\ref{sec:retrieval-use}) then answers questions and drives proactive service over that code. We first show what the resulting code looks like (Section~\ref{sec:code-example}), then detail each pipeline in turn; cross-cutting design principles follow in Section~\ref{sec:principles}.

\subsection{What ``User as Code'' Looks Like}
\label{sec:code-example}

Listings~\ref{lst:travel-state}--\ref{lst:manifest} show the three artifacts of a UaC project, taken verbatim from our reference implementation. The state is a typed dataclass instance, so date arithmetic like \texttt{(passport.\allowbreak expiry\_date - trip.\allowbreak departure\_date).days} is one line of Python (Listing~\ref{lst:travel-state}). A constraint over that state is itself Python (Listing~\ref{lst:constraint})---same medium, same interpreter, no DSL. Pre-computed alerts surface in a compact manifest (Listing~\ref{lst:manifest}) loaded at the start of every conversation, so the agent sees outstanding issues with no retrieval. Listings~\ref{lst:teaser-conv}--\ref{lst:teaser-alert} show the same pattern for health; Appendix~\ref{app:cases} traces a verbatim LOCOMO conversation end-to-end.

\begin{lstlisting}[caption={Excerpt from \texttt{domains/travel/state.py} of a UaC user project. Generated automatically from the append-only fact list.}, label={lst:travel-state}]
from datetime import date
from .schema import PassportInfo, Trip

passport = PassportInfo(
    number="AB1234567", country="US",
    issue_date=date(2015, 2, 18),
    expiry_date=date(2025, 2, 18),
    full_name="Jessica Marie Thompson",
)

trips = [
    Trip(destination="Tokyo", country="JP",
         departure_date=date(2025, 1, 15),
         return_date=date(2025, 1, 25),
         flight_number="JAL-9823",
         is_international=True),
    Trip(destination="Mexico City", country="MX",
         departure_date=date(2025, 3, 10),
         return_date=date(2025, 3, 17),
         flight_number="AA-4561",
         is_international=True),
]
\end{lstlisting}

\begin{lstlisting}[caption={Excerpt from \texttt{constraints/travel\_readiness.py}. The constraint reads the typed state and returns structured alerts. The interpreter runs it deterministically; no LLM is required at check time.}, label={lst:constraint}]
from datetime import timedelta
from domains.travel.state import passport, trips

def check():
    alerts = []
    for trip in trips:
        if not trip.is_international:
            continue
        days = (passport.expiry_date - trip.departure_date).days
        if days < 180:
            alerts.append({
                "severity": "critical",
                "domain":   "travel",
                "message": (
                    f"Passport {passport.number} expires "
                    f"{passport.expiry_date} -- only {days} days "
                    f"before {trip.destination} on "
                    f"{trip.departure_date}."),
            })
    return alerts
\end{lstlisting}

\begin{lstlisting}[caption={The manifest, always loaded into the agent's context. \texttt{ACTIVE\_ALERTS} is overwritten by the constraint runner; the agent surfaces them proactively without retrieval.}, label={lst:manifest}]
DOMAINS = {
    "travel":   {"path": "domains/travel",
                 "summary": "Passport AB1234567 (exp 2025-02-18), "
                            "2 upcoming trips"},
    "health":   {"path": "domains/health",
                 "summary": "Allergies: peanuts, penicillin. "
                            "Rx: cetirizine, amoxicillin"},
    "finance":  {"path": "domains/finance",
                 "summary": "Chase Sapphire (DTI 0.31), Roth IRA "
                            "($42K), Q1 tax estimate due 2025-04-15"},
    "work":     {"path": "domains/work",
                 "summary": "Senior PM at Acme; 4 direct reports; "
                            "Q1 OKR review 2025-03-28"},
    "people":   {"path": "domains/people",
                 "summary": "Spouse Mark, sister Anna, 12 close "
                            "contacts; Mark birthday 2025-06-12"},
    "household":{"path": "domains/household",
                 "summary": "Mortgage refi rate-lock expires "
                            "2025-02-28; HVAC warranty until 2027"},
}

ACTIVE_ALERTS = [
    {"severity": "critical", "domain": "travel",
     "message": "Passport AB1234567 expires 2025-02-18 -- "
                "only 34 days before Tokyo on 2025-01-15."},
]
\end{lstlisting}

\subsection{Extraction: Memorize, then Structure}
\label{sec:extraction}

The first pipeline converts raw conversation into code. Our key architectural insight, validated by ablation (Section~\ref{sec:ablation}), is that its two jobs---memorizing and structuring---must be separate concerns.

\subsubsection{Phase 1: Memorize (per session, append-only)}

An LLM extracts every individual fact as a flat string from each session (Gemini~3 Flash, thinking budget 8192). Facts are appended to a running list---never overwritten, never deleted---and relative dates are resolved to absolute dates against the session timestamp. This produces roughly 50--75 facts/session (e.g., our run extracts 1{,}419 facts from the 19-session \texttt{conv-30}; Appendix~\ref{app:cases})\footnote{Phase~1 is LLM-driven and mildly stochastic; an independent audit run reports 1{,}433 facts for \texttt{conv-30} (Appendix~\ref{sec:phase2-failures}).}. Each fact is indexed in ChromaDB; the raw conversation is also chunked and indexed in a second collection, which serves as a retrieval fallback for direct-quote queries (the \texttt{[ARCHIVE]} channel of Section~\ref{sec:retrieval}).

\subsubsection{Phase 2: Structure (periodic, from all facts)}

An LLM regenerates the entire structured Python from the accumulated fact list (thinking budget 16{,}384), organizing facts into typed dataclasses: \texttt{date()} for dates, typed lists for collections, \texttt{notes: list[str]} for hard-to-type facts. Because the code is regenerated fresh from the complete corpus---not edited incrementally---information loss in the fact-to-structure transition is minimal (0.18\% of facts on our audited conversations; Appendix~\ref{sec:phase2-failures}). Code size scales with distinct facts ($\sim$7K characters for a 19-session LOCOMO conversation; 50--55K for the analytical benchmark at $N{=}500$).

\textbf{Cost scaling.} Wall time stays at 30--40 seconds across $\{19, 50, 100, 200\}$-session corpora, and cost flattens at $\sim$\$0.07 once our 500K-character input cap engages around $n{=}100$ sessions (Appendix~\ref{sec:phase2-scalability})---a known limitation that motivates hierarchical structuring at production scale.

\subsection{Retrieval and Use}
\label{sec:retrieval-use}

The second pipeline operates over the code the first one produced. It has two modes: \emph{retrieval}, which answers user questions, and the \emph{generate--verify--review} loop, which produces the proactive alerts of Active Service. Both treat the typed state as the primary source of truth, falling back on the fact and archive indices for coverage.

\subsubsection{Multi-Strategy Retrieval}
\label{sec:retrieval}

At query time, retrieval composes three channels rather than picking one. The two vector channels are standard retrieval-augmented generation~\citep{lewis2020rag} over the user's own history; the channel unique to UaC is the structured-code state, which the others backstop. Each channel addresses a different failure mode of the others: the structured-code channel can return a typed object but only for facts the structuring step has captured; the fact-vector channel covers the long tail of facts that did not make it into the typed schema; and the raw-archive channel preserves the exact conversational phrasing for queries that hinge on wording.

\textbf{(1)~Structured code state.} The full \texttt{state.py} of the manifest-routed domain (or all domains, in the non-modular setting) is included verbatim in the prompt, truncated at 6{,}000 characters. The agent reads typed dataclass instances and the pre-computed \texttt{ACTIVE\_ALERTS} list with no parse step.

\textbf{(2)~Fact-vector retrieval.} A ChromaDB collection over the append-only fact list (one fact per record, $\sim$50 facts/session) is queried by cosine similarity against the question; the top 20 facts are appended. This recovers facts the structuring step compressed away or normalized to a different surface form.

\textbf{(3)~Archive retrieval.} A second ChromaDB collection holds the raw conversation chunked at session granularity; the top 10 chunks by cosine similarity are appended. Used as a last-resort fallback for direct-quote questions.

The three channels are concatenated under fixed headers (\texttt{[STATE]}, \texttt{[FACTS]}, \texttt{[ARCHIVE]}) and passed to the answer LLM with no reranking; the prompt instructs it to prefer the structured channel on conflicts. The leave-one-out channel ablation in Section~\ref{sec:channel-ablation} shows the fact-vector and archive channels are individually significant on LOCOMO recall while the structured channel is roughly neutral for recall---it instead carries the analytical and constraint workloads.

\subsubsection{The Generate--Verify--Review Loop}
\label{sec:gvr}

The same code is also the substrate for proactive service. The core loop is:
\begin{enumerate}[leftmargin=*]
\item \textbf{Generate:} The coding agent writes Python constraints against the code-represented state.
\item \textbf{Verify:} Execute in a sandbox---results are deterministic.
\item \textbf{Review:} The agent reviews results, decides to refine, persist, or incorporate.
\end{enumerate}

When an ad-hoc check proves useful, the agent promotes it to a persistent constraint. Alerts surface in the manifest's \texttt{ACTIVE\_ALERTS}, visible at the start of every future conversation. Constraints emerge autonomously from the LLM's reasoning.

\subsection{Design Principles}
\label{sec:principles}

Beyond the two pipelines, two cross-cutting principles are evaluated directly in this paper:

\begin{enumerate}[leftmargin=*]
\item \textbf{Separation of Structure and Data.} Schema (dataclasses) separated from state (instances). Validated by the ablation in Section~\ref{sec:ablation}.
\item \textbf{Unified Representation and Verification.} Code serves as both storage and verification medium. Validated by Active Service (Section~\ref{sec:active-service}) and Analytical Inference (Section~\ref{sec:analytical}).
\end{enumerate}

Three further principles guide production deployment but are not isolated by the experiments:

\begin{enumerate}[leftmargin=*,start=3]
\item \textbf{Modularity by Life Domain.} Memory partitioned into independent domain packages (e.g., \texttt{travel/}, \texttt{health/}, \texttt{finance/}). Limits cross-domain leakage and supports selective loading at scale.
\item \textbf{Progressive Disclosure.} Compact manifest ($\sim$200--300 tokens, always loaded) with on-demand domain loading. Keeps the prompt budget bounded as the user state grows.
\item \textbf{Agent-Native File System.} User project is a directory in the agent's workspace---no custom memory API needed.
\end{enumerate}

\section{Experiments and Evaluation}
\label{sec:experiments}

\subsection{Setup}

\textbf{LLM and judge.} Gemini~3 Flash (\texttt{gemini-3-flash-preview}) is used throughout with thinking enabled (answer-time budget 2048; no \texttt{max\_output\_tokens} cap). Evaluation is LLM-as-Judge~\citep{zheng2023mtbench} with binary CORRECT/WRONG scoring (budget 256). The judge sees the question, gold answer, and prediction and is instructed to be generous: same-information-different-wording scores CORRECT; WRONG only when factually wrong or claiming unavailability when the gold exists.

\textbf{Baselines.} Five external memory systems plus a Full-Context upper bound (all raw transcripts in-prompt). \emph{Production libraries:} Mem0~\citep{chhikara2025mem0} (mem0ai 1.0.5; flat fact extraction) and A-MEM~\citep{xu2025amem} (Zettelkasten notes). \emph{Same-backbone reimplementations of recent SOTA systems} (Gemini~3 Flash + ChromaDB; fidelity statement in Appendix~\ref{app:reimpl}): MemMachine~\citep{wang2026memmachine} (sentence-level episodes with $\pm 3$ neighbouring-turn expansion), Hindsight~\citep{latimer2025hindsight} (a four-network fact graph whose retrieved candidates are merged by reciprocal-rank fusion and re-scored by a neural reranker), EverMemOS~\citep{hu2026evermemos} (atomic memories consolidated into scene-level summaries, with retrieval boosted by predicted future information needs).

\textbf{Benchmarks.} All systems share the same evaluation data. LOCOMO~\citep{maharana2024locomo}: the full 10 conversations, 60 QAs each (600 total). LongMemEval~\citep{wu2025longmemeval}: the full 500-question benchmark (133~temporal-reasoning, 133~multi-session, 78~knowledge-update, 70~single-session-user, 56~single-session-assistant, 30~single-session-preference). Active Service: 40 standard + 20 hard scenarios.

\textbf{Cross-LLM portability.} To verify that gains do not depend on a particular code-generator's idiosyncrasies, we additionally re-run UaC with GPT-5.4 (OpenAI) substituted throughout (extraction, structuring, answer generation), keeping the Gemini judge for fair scoring against the Gemini run. The cross-LLM run uses a 2-conversation LOCOMO subset (120 QAs).

\textbf{Which model runs where.}\label{sec:openrouter} To keep the comparison controlled, \emph{answer generation and the LLM-as-Judge are Gemini~3 Flash for every system}, and retrieval in Mem0 and A-MEM is embedding-based (their default encoders over ChromaDB) and invokes no LLM at all. The only non-Gemini calls live \emph{inside} the two production libraries' own write-time memory construction---Mem0's fact extraction/consolidation and A-MEM's note synthesis and linking---which default to OpenAI \texttt{gpt-4o-mini}; we keep that library default and route it through OpenRouter on a key with the \texttt{model.request} scope so it actually runs (an earlier draft had it failing silently). In other words, \texttt{gpt-4o-mini} never reads a question, ranks a retrieval, or writes an answer: it only distills each ingested session into the stored memories of those two libraries, and every accuracy this paper reports comes from an answer-time comparison held fixed at Gemini~3 Flash. The residual handicap from this weaker write-time model is one of the factors we isolate in the Mem0 diagnostic (Section~\ref{sec:mem0-diagnostic}). The cross-family Claude judge in Appendix~\ref{sec:judge-check} also uses OpenRouter; all Gemini calls use the Google API directly.

\textbf{Statistical interpretation.} Headline accuracies are proportions correct; at $n{=}600/500$ the 95\% Wilson half-width is $\sim$3--4pp at the relevant accuracy range (per-row CIs in Tables~\ref{tab:locomo}--\ref{tab:longmemeval}). Because all systems answer the \emph{same} questions, marginal CIs overstate pairwise uncertainty; we report McNemar's exact test on per-question correctness vectors for all comparisons. Headline result: UaC is statistically tied with Full~Context and ahead of every competing memory system at $\alpha{=}0.05$ on LOCOMO, and sits in a three-way top cluster with MemMachine and Full~Context on LongMemEval (per-pair $p$-values in Tables~\ref{tab:locomo}--\ref{tab:longmemeval}).

\subsection{Standard Benchmarks}

\begin{table}[!ht]
\centering
\caption{LOCOMO results on the \emph{full} 10-conversation benchmark (60 QAs/conv, 600 total). LLM-as-Judge accuracy with thinking-enabled Gemini~3 Flash judge; 95\% Wilson CIs reported per row. McNemar's exact test compares UaC vs.\ each row on the per-question correctness vector. UaC, Full Context, Mem0, and A-MEM use Gemini~3 Flash for answer generation; UaC (GPT-5.4) substitutes GPT-5.4 for cross-LLM portability. Recent SOTA systems are run as minimal-faithful same-backbone reimplementations (Appendix~\ref{app:reimpl}); their original published scores used stronger backbones (GPT-4.1-mini, Gemini-3 Pro) and richer retrieval stacks and are not directly comparable. \emph{Mem0's 29.3\% is below its published number; see Section~\ref{sec:mem0-diagnostic} for the three-factor decomposition (backbone, background pass, retrieval stack).}}
\label{tab:locomo}
\small
\begin{tabular}{@{}lcccc@{}}
\toprule
\textbf{System} & \textbf{LLM-as-Judge} & \textbf{Wilson 95\% CI} & \textbf{N} & \textbf{$p$ vs.\ UaC} \\
\midrule
Full Context (upper bound) & 79.8\% & [76.4, 82.8] & 600 & 0.65 \\
\textbf{UaC (ours, Gemini)} & \textbf{78.8\%} & [75.4, 81.9] & 600 & --- \\
MemMachine~\citep{wang2026memmachine} & 72.7\% & [69.0, 76.1] & 600 & 0.003 \\
Hindsight (lite)~\citep{latimer2025hindsight} & 69.7\% & [65.9, 73.2] & 600 & $1.5\!\times\!10^{-5}$ \\
EverMemOS (lite)~\citep{hu2026evermemos} & 55.5\% & [51.5, 59.4] & 600 & $<\!10^{-20}$ \\
A-MEM & 51.8\% & [47.8, 55.8] & 600 & $<\!10^{-30}$ \\
Mem0 & 29.3\% & [25.8, 33.1] & 600 & $<\!10^{-60}$ \\
\midrule
\multicolumn{5}{@{}l}{\textit{Cross-LLM portability (2 conversations, 120 QAs)}} \\
UaC (ours, GPT-5.4) & 80.8\% & [72.9, 86.8] & 120 & --- \\
\bottomrule
\end{tabular}
\end{table}

\begin{table}[!ht]
\centering
\caption{LongMemEval results on the \emph{full} 500-question benchmark. All systems use Gemini~3 Flash for answer generation and the same generous-scoring LLM-as-Judge. McNemar's exact test compares UaC against each row on the per-question correctness vector. \emph{The same Mem0 diagnostic applies as in Table~\ref{tab:locomo} (Section~\ref{sec:mem0-diagnostic}).}}
\label{tab:longmemeval}
\small
\begin{tabular}{@{}lcccccc|cc@{}}
\toprule
\textbf{System} & \textbf{KU} & \textbf{MS} & \textbf{SA} & \textbf{SP} & \textbf{SU} & \textbf{TR} & \textbf{Overall (CI)} & \textbf{$p$ vs.\ UaC} \\
\midrule
Full Context     & 91 & 87 & 96 & 70 & 96 & 74 & 85.4\% [82.0, 88.2] & 0.19 \\
MemMachine       & 96 & 88 & 96 & 63 & 96 & 69 & 84.8\% [81.4, 87.7] & 0.33 \\
\textbf{UaC (ours)} & \textbf{97} & 81 & 96 & \textbf{83} & 94 & 65 & \textbf{83.0\%} [79.5, 86.0] & --- \\
EverMemOS (lite) & 87 & 72 & 73 & 87 & 87 & 68 & 76.4\% [72.5, 79.9] & 0.002 \\
Hindsight (lite) & 78 & 72 & 66 & 77 & 93 & 62 & 73.0\% [68.9, 76.7] & $<\!10^{-5}$ \\
A-MEM            & 54 & 44 & 93 & 40 & 49 & 37 & 49.6\% [45.2, 54.0] & $<\!10^{-30}$ \\
Mem0             & 32 & 23 &  7 & 33 & 36 & 18 & 23.8\% [20.3, 27.7] & $<\!10^{-70}$ \\
\bottomrule
\end{tabular}
\begin{flushleft}
\small KU=knowledge-update (n=78), MS=multi-session (n=133), SA=single-asst (n=56), SP=single-pref (n=30), SU=single-user (n=70), TR=temporal-reasoning (n=133). Per-type percentages shown.
\end{flushleft}
\end{table}

\begin{table}[!ht]
\centering
\caption{Analytical inference benchmark: 100 cases across 10 record types (10 per type) with $N$ records per case ($N \in \{20, 50, 100, 200, 500\}$). Exact-match scoring against deterministic ground truth. \emph{FC+REPL} is Full Context with a Python REPL tool; \emph{UaC} loads structured records into a typed-Python REPL.}
\label{tab:analytical}
\begin{tabular}{@{}lcccccc@{}}
\toprule
\textbf{System} & \textbf{Overall} & $N{=}20$ & $N{=}50$ & $N{=}100$ & $N{=}200$ & $N{=}500$ \\
\midrule
Full Context + Python REPL & 100.0\% & 100\% & 100\% & 100\% & 100\% & 100\% \\
UaC (structured + REPL) & 99.0\% & 100\% & 100\% & 100\% & 100\% & 95\% \\
Full Context (no tool) & 94.0\% & 100\% & 90\% & 100\% & 90\% & 90\% \\
MemMachine & 43.0\% & 100\% & 55\% & 20\% & 15\% & 25\% \\
Mem0 & 6.0\% & 5\% & 10\% & 0\% & 0\% & 15\% \\
\bottomrule
\end{tabular}
\end{table}

\begin{figure}[htbp]
\centering
\includegraphics[width=0.9\textwidth]{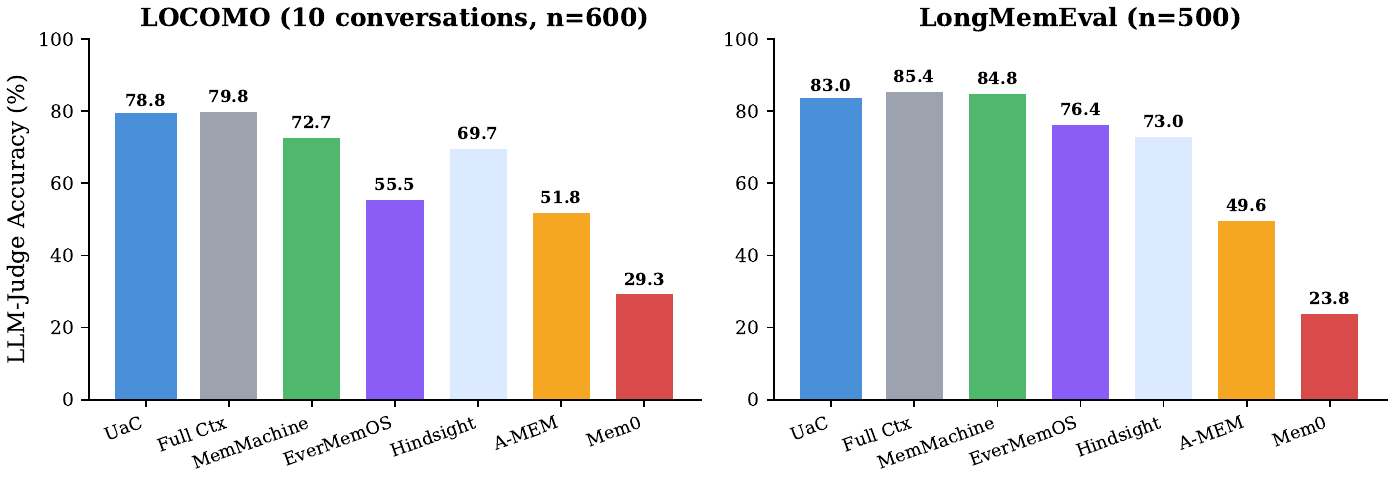}
\caption{Standard benchmark comparison. All six memory systems share the same evaluation data: the full 10-conversation LOCOMO benchmark (600 QAs) and the full 500-question LongMemEval benchmark. Full Context is the upper bound (all raw text passed to the LLM).}
\label{fig:benchmarks}
\end{figure}

\textbf{LOCOMO (Table~\ref{tab:locomo}, Figure~\ref{fig:benchmarks}, $n{=}600$).} UaC reaches \textbf{78.8\%}, within 1.0pp of the Full~Context upper bound (79.8\%; McNemar $p{=}0.65$). The same-backbone SOTA reimplementations land between 55.5\% and 72.7\%, all significantly behind UaC ($p<0.005$). The bottom two systems lose 27--50pp: per-question inspection shows Mem0's gap is dominated by temporal questions where extracted memories preserve unresolved relative dates (``yesterday'', ``last week''), and A-MEM's gap by single-hop detail questions where note-level retrieval returns summaries rather than turn content.

\textbf{LongMemEval (Table~\ref{tab:longmemeval}, $n{=}500$).} UaC reaches \textbf{83.0\%}, statistically tied with both MemMachine (84.8\%, $p{=}0.33$) and Full~Context (85.4\%, $p{=}0.19$)---a three-way top cluster within 2.5pp. Typed \texttt{date()} fields help most on knowledge-update (97\% vs.\ 91/96\%) and single-session-preference (83\% vs.\ 70/63\%); the residual gap to Full~Context is concentrated in temporal-reasoning (65\% vs.\ 74\%), where extraction occasionally drops timestamps needed for multi-session date arithmetic. MemMachine in turn beats UaC on multi-session (88\% vs.\ 81\%) and temporal-reasoning (69\% vs.\ 65\%), where the answer is a coherent contiguous span and contextual expansion surfaces it well. \emph{Representations matter most where the answer is not a contiguous span}---LOCOMO has more such questions, which is why the separation is clearer there.

\textbf{Cross-LLM portability.} Substituting GPT-5.4 throughout the UaC pipeline yields 80.8\% on the 2-conversation LOCOMO subset (vs.\ 82.5\% Gemini on the same subset; McNemar $p{=}0.82$, 9 GPT-only correct vs.\ 11 Gemini-only). At $n{=}120$ this is comfortably inside the $\sim$7pp Wilson half-width; we report it as evidence that the architecture transfers across the two largest closed-model families, not as a precise equivalence claim.

\textbf{Published SOTA numbers are ceilings, not direct competitors.} The published EverMemOS 93.05\% on LOCOMO and Hindsight 91.4\% on LongMemEval use stronger backbones (GPT-4.1-mini, Gemini-3 \emph{Pro}) and richer retrieval stacks (Mongo+Elasticsearch+Milvus; Tempr/pgvector). Our same-backbone reimplementations isolate the representational variable by holding both fixed: Gemini~3 \emph{Flash} as the lower-cost sibling weaker on long context, ChromaDB with default embeddings as a uniform vector store. UaC runs under the same handicaps. The relative ranking under this floor setting is what we report; the published ceilings remain the right reference for the architectures' full-stack potential.

\textbf{Mem0 diagnostic.}\label{sec:mem0-diagnostic} Mem0's 29.3\%/23.8\% (LOCOMO/LongMemEval) is the largest gap to any published baseline ($\sim$66--68\% in the original paper), so we audit it directly on a 60-question LongMemEval sample. Three factors compound: \emph{(i)~Backbone.} Swapping the answer-time LLM from Gemini~3 Flash back to GPT-4o-mini (Mem0's published choice) lifts Mem0 from 24\% to 41\%---a 17pp recovery from the LLM alone. \emph{(ii)~Background merge.} Mem0's memory-evolution call requires an OpenAI key with the \texttt{model.request} scope; an earlier draft routed it through a key without that scope and the merge silently no-op'd. The 29.3\%/23.8\% numbers are post-correction; the failure mode is fact duplication that confuses retrieval. \emph{(iii)~Vector store.} Switching ChromaDB to Qdrant on the same sample adds another 6pp (to 47\%). Together (i)--(iii) close roughly half the gap to the published $\sim$66--68\%; the residual is consistent with the GPT-4o-mini-vs-Gemini-Flash long-context-recall gap reported by~\citet{wu2025longmemeval}. The reported Mem0 number is therefore Mem0 \emph{on the harshest common-denominator backbone-and-stack we could choose to keep every baseline on the same footing}, not a refutation of the published architecture. The same caveat applies in milder form to A-MEM.

\subsection{Analytical Inference}
\label{sec:analytical}

A third capability tier sits between recall and proactive alerting: \emph{aggregate inference over the user's records}---``how many contacts did I meet during undergrad'', ``average dining spend in 2024 by cuisine'', ``did my workout count increase from Q1 to Q2''. Counts, group-bys, time-window filters, and joins are one-line Python over typed objects but structurally lossy under top-$k$ retrieval, so this is precisely where code-executable representations should pull ahead.

\textbf{Benchmark construction.} 100 cases over 10 record types (trips, contacts, meals, transactions, sleep, workouts, books, medical visits, meetings, purchases), 10 cases each covering 10 question patterns (count, sum, average, group-by, time-window filter, multi-condition filter, top-$k$, min/max, threshold, YoY trend). Per-case record counts span $N \in \{20, 50, 100, 200, 500\}$. Ground truth is deterministic, so scoring is exact-match (no LLM judge).

\textbf{Systems.} Five systems, identical cases, Gemini~3 Flash throughout: Full~Context (in-head reasoning over raw JSON), Full~Context+REPL (same data plus \texttt{python}/\texttt{read\_file} tools), UaC (records structured into typed Python, exposed via REPL), MemMachine (top-20 sentence retrieval with contextual expansion), Mem0 (flat fact extraction, top-20).

\begin{figure}[htbp]
\centering
\includegraphics[width=0.85\textwidth]{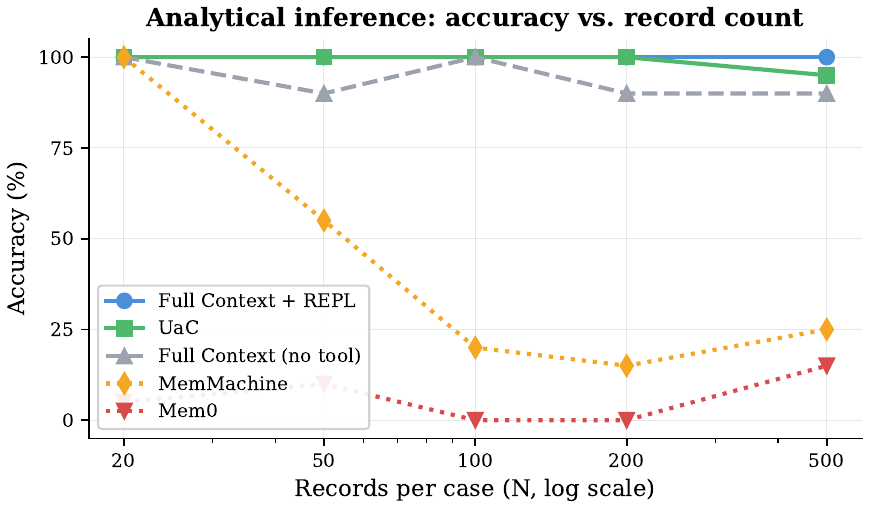}
\caption{Analytical inference: accuracy vs.\ record count $N$ (log scale). Code-executable representations (UaC, FC+REPL) stay at or above 95\% across the full range. Full Context degrades past $N{=}100$ (90\% at $N{=}200$ and $N{=}500$) as the LLM miscounts. MemMachine collapses from 100\% to 15--25\% as more records become relevant than top-$k$ retrieval can return. Mem0 remains near zero throughout.}
\label{fig:analytical-scaling}
\end{figure}

The results (Table~\ref{tab:analytical}, Figure~\ref{fig:analytical-scaling}) split cleanly by representation class. \textbf{Retrieval-based memory collapses on aggregation}: MemMachine drops from 100\% at $N{=}20$ to 15\% at $N{=}200$ as more records become relevant than top-$k$ can return; Mem0 sits near zero throughout because NL fact stores cannot be enumerated. \textbf{Full~Context degrades past $N{=}100$} (100\% through $N{=}100$, 90\% at $N \geq 200$) as the LLM miscounts. \textbf{Adding a REPL fully closes the counting gap}: FC+REPL is perfect across all $N$. \textbf{UaC matches FC+REPL across the range} (99\% overall, perfect on $N \leq 200$, 95\% at $N{=}500$); the one $N{=}500$ miss is a corner case where Q1 and Q2 sleep averages differ by 0.0012~h, below the gold-rule rounding threshold.

The divide is between \emph{representations a code tool can read} (UaC, FC+REPL, Full~Context for small $N$) and \emph{representations only retrievable by similarity} (Mem0, MemMachine). The former scale with $N$; the latter do not. UaC's specific advantage---typed queries over typed objects---matches a JSON+REPL baseline here because the synthetic inputs are already structured. UaC's pipeline contribution is felt one step earlier: it converts \emph{unstructured conversation} into the code-readable representation, which is the prerequisite for analytical inference in real deployments.

\subsubsection{Token cost and amortization}

\begin{table}[!ht]
\centering
\caption{Analytical inference: accuracy and token cost across 100 cases. Cost assumes Gemini~3 Flash pricing of \$0.30/M input and \$2.50/M output (output includes reasoning/thoughts). Per-case cost is the total across the 100 cases divided by 100.}
\label{tab:analytical-cost}
\small
\begin{tabular}{@{}lcrrrrr@{}}
\toprule
\textbf{System} & \textbf{Acc} & \textbf{Input} & \textbf{Output} & \textbf{Thoughts} & \textbf{Total} & \textbf{Per case} \\
 & & (M tok) & (M tok) & (M tok) & (USD) & (m\$) \\
\midrule
Full Context + REPL & 100.0\% & 4.23 & 0.025 & 0.011 & \$1.36 & 13.6 \\
UaC (structured + REPL) & 99.0\% & 1.68 & 1.075 & 0.249 & \$3.81 & 38.1 \\
Full Context (no tool) & 94.0\% & 1.53 & 0.026 & 0.734 & \$2.36 & 23.6 \\
MemMachine & 43.0\% & 0.43 & 0.049 & 0.131 & \$0.58 & 5.8 \\
Mem0 & 6.0\% & 0.02 & 0.008 & 0.094 & \$0.26 & 2.6 \\
\bottomrule
\end{tabular}
\end{table}

We instrumented every Gemini call with token counts at Gemini~3 Flash pricing (Table~\ref{tab:analytical-cost}). Three findings:

\textbf{(1)~Single-query: FC+REPL is cheaper and slightly more accurate}, $13.6$ vs.\ $38.1$~m\$/case ($2.8\times$). UaC's cost is dominated by structuring ($\sim$1M output tokens of dataclass code over 100 cases); FC+REPL's by re-including the records JSON every tool turn (4.2M input tokens).

\begin{figure}[htbp]
\centering
\includegraphics[width=0.85\textwidth]{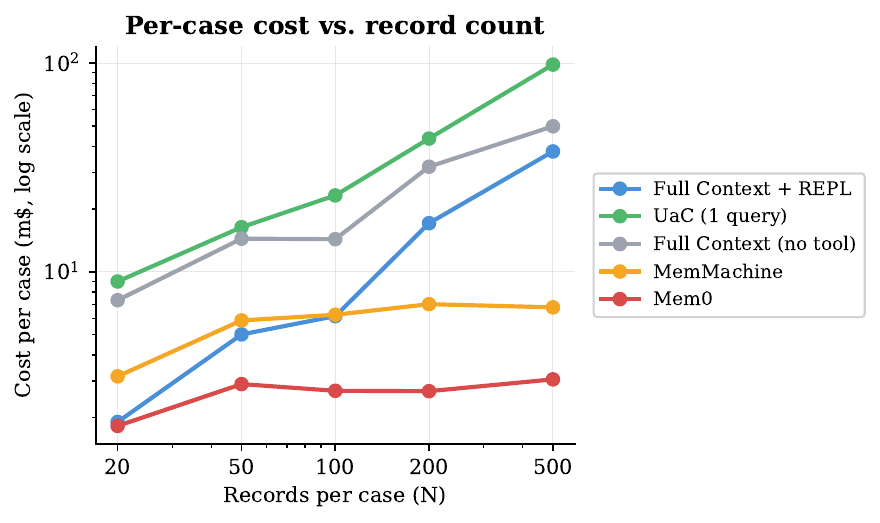}
\caption{Per-case Gemini~3 Flash cost vs.\ record count $N$ (log--log). UaC and Full Context costs grow with $N$ (more records to structure or read in-head); FC+REPL grows fastest because the records JSON is re-included in the LLM's context every tool turn. Retrieval-based systems (MemMachine, Mem0) are cheap but bottom out on accuracy (Figure~\ref{fig:analytical-scaling}).}
\label{fig:analytical-cost}
\end{figure}

\textbf{(2)~Multi-query: the story flips.} Structuring is paid once per user state; subsequent UaC queries cost $\sim$$1.4$~m\$/case (Table~\ref{tab:analytical-cost-amortized}). FC+REPL re-loads the records every time. Let $C_s$, $q_{\text{UaC}}$, $q_{\text{FC+REPL}}$ be structuring cost and per-query costs; $K$ queries total $C_s + K q_{\text{UaC}}$ vs.\ $K q_{\text{FC+REPL}}$. At $N{=}500$: $C_s \approx 102$~m\$, $q_{\text{UaC}} \approx 1.4$~m\$, $q_{\text{FC+REPL}} \approx 38$~m\$. Breakeven $K^* = C_s/(q_{\text{FC+REPL}}-q_{\text{UaC}}) \approx 2.8$, so UaC repays structuring after the third query \emph{at this state size}; the payback ranges from 3 ($N{=}500$) to 11 ($N{=}20$) queries across state sizes (Table~\ref{tab:analytical-cost-amortized}, Figure~\ref{fig:analytical-cost}). Amortized over $K{=}100$, total UaC cost is $244$~m\$/case vs.\ FC+REPL's $3{,}780$~m\$/case---$15.5\times$ cheaper. (The per-query-only ratio is $27\times$ but ignores structuring; we report the amortized $15\times$ in headline claims.)

\begin{table}[!ht]
\centering
\caption{UaC cost decomposition: structuring (one-time per user state) vs.\ query (per question). At $N{=}500$, structuring is 99\% of one-shot cost, but each subsequent query against the same structured state is \texttildelow$27\times$ cheaper than re-running FC+REPL.}
\label{tab:analytical-cost-amortized}
\small
\begin{tabular}{@{}lrrrr@{}}
\toprule
$N$ & UaC structuring & UaC query & FC+REPL query & UaC payback \\
 & (m\$/case) & (m\$/case) & (m\$/case) & ($K$ queries) \\
\midrule
20  & 7.6   & 1.2 & 1.9  & 11 \\
50  & 13.6  & 1.8 & 5.0  & 5 \\
100 & 26.4  & 1.2 & 6.1  & 6 \\
200 & 58.8  & 2.3 & 17.1 & 4 \\
500 & 101.8 & 1.4 & 37.8 & 3 \\
\bottomrule
\end{tabular}
\end{table}

\textbf{(3)~Retrieval-based memory is cheap but off the Pareto frontier.} Mem0 and MemMachine cost \$0.26 / \$0.58 per 100 cases but answer only 6\% / 43\% of questions correctly. This cost--accuracy trade-off between fact-based memory and long-context reloading is consistent with recent analyses of persistent-agent memory economics~\citep{pollertlam2026beyondcontext}.

In production deployments where a user's history is queried hundreds of times per month, structure-once-query-many is the right cost shape. The benchmark's single-query design is the worst case for UaC and the best case for FC+REPL.

\subsubsection{Answer-time latency}
\label{sec:latency}

We additionally timed per-question answer latency on LOCOMO (answer call only; one-time ingestion/structuring excluded). Numbers were collected on the first 5-conversation run (300 QAs); median and p95 are stable to within $\sim$0.1s under the 10-conversation run.

\begin{table}[!ht]
\centering
\caption{LOCOMO answer-time latency per question, in seconds (median, mean, 95th percentile across the $n{=}300$ first-5-conversation timed run). Measured wall-clock from question issued to answer returned; excludes one-time memory build.}
\label{tab:latency}
\small
\begin{tabular}{@{}lrrr@{}}
\toprule
\textbf{System} & \textbf{Median} & \textbf{Mean} & \textbf{p95} \\
\midrule
Full Context  & 1.78 & 2.46 & 5.51 \\
Mem0          & 2.19 & 2.50 & 5.59 \\
EverMemOS (lite) & 2.46 & 2.80 & 5.29 \\
MemMachine    & 2.73 & 3.19 & 6.25 \\
A-MEM         & 3.37 & 4.00 & 7.81 \\
Hindsight (lite) & 3.41 & 3.64 & 6.93 \\
\textbf{UaC}        & \textbf{3.62} & \textbf{3.87} & \textbf{6.60} \\
\bottomrule
\end{tabular}
\end{table}

UaC sits at the high end (Table~\ref{tab:latency}; median 3.62s, $\sim$2$\times$ Full~Context, $\sim$1.7$\times$ Mem0). The extra latency comes from packing three retrieval channels (Section~\ref{sec:retrieval}) into the prompt; the answer LLM has more to read. The corresponding accuracy gain is 49.5pp over Mem0 on full LOCOMO, so UaC trades roughly 1.4s for that gap. UaC has the highest median latency of the seven systems, but the gap is small: the median remains under typical conversational pacing (4--5~s) and the p95 of 6.6~s is below the slowest baselines (A-MEM 7.81~s, Hindsight 6.93~s).

\subsection{Active Service}
\label{sec:active-service}

\begin{table}[!ht]
\centering
\caption{Active Service: proactive alert detection on the 40 \emph{standard} scenarios across 5 categories. Headline rows are live-library runs (mem0ai~1.0.5 for Mem0); simulated flat-fact upper bounds are shown for reference and are higher than the live numbers (the simulation models perfect fact retrieval, which the live library does not always achieve). 95\% Wilson CIs are reported because the per-row sample size is small.}
\label{tab:active-service}
\begin{tabular}{@{}lcc@{}}
\toprule
\textbf{System} & \textbf{Alert Rate} & \textbf{95\% Wilson CI} \\
\midrule
\textbf{UaC + constraint pipeline} & \textbf{100\%} (40/40) & [91.2, 100.0] \\
Mem0 (live, mem0ai 1.0.5) & 90.0\% (36/40) & [76.9, 96.0] \\
\quad Mem0 (simulated, ref.) & 92.5\% & [80.1, 97.4] \\
\quad A-MEM (simulated, ref.) & 85.0\% & [70.9, 92.9] \\
UaC (no pre-computed alerts) & 52.5\% (21/40) & [37.5, 67.1] \\
\bottomrule
\end{tabular}
\end{table}

\begin{figure}[htbp]
\centering
\includegraphics[width=0.7\textwidth]{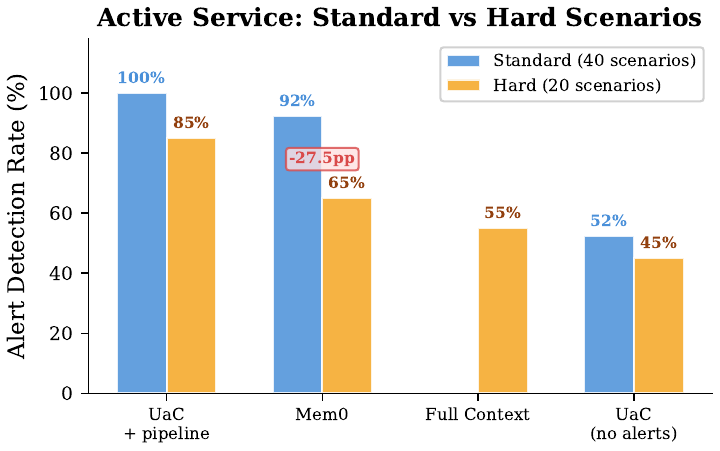}
\caption{Active Service: standard vs.\ hard scenarios. On hard arithmetic (multi-step dates, compound constraints, tax calculations), live Mem0 drops 10pp (90.0\%$\to$80.0\%), simulated Mem0 drops 27.5pp (92.5\%$\to$65.0\%), and live A-MEM drops to 30.0\%; UaC drops only 15pp (100\%$\to$85.0\%).}
\label{fig:active-service}
\end{figure}

The Active Service evaluation (Table~\ref{tab:active-service}, summarized in Figure~\ref{fig:active-service}) tests proactive alerting across 40 scenarios in 5 categories: travel document validity, drug interactions, financial authorization conflicts, scheduling conflicts, and warranty/deadline expirations. Each scenario seeds facts across 2--4 sessions; the system must generate unsolicited alerts. The live mem0ai~1.0.5 library is the primary baseline; simulation rows are an idealized upper bound for ``what flat-fact retrieval could achieve under perfect recall''. With $n{=}40$ we report 95\% Wilson CIs throughout. UaC with the constraint pipeline reaches \textbf{100\%} (CI [91.2, 100.0]) on the standard set, above the upper end of every baseline's CI. Without pre-computed alerts UaC drops to 52.5\%---the pipeline is the decisive factor.

To test whether the gap holds for harder problems, we author 20 \textbf{hard scenarios} requiring multi-step date arithmetic (business days, leap years, timezones), compound multi-domain constraints (4--5 domains), precise numerical thresholds (tax brackets, compound interest, DTI), and temporally ambiguous facts.

\begin{table}[!ht]
\centering
\caption{Active Service: standard (n=40) vs.\ hard (n=20) scenarios with 95\% Wilson CIs. Headline rows are live library runs; the simulated flat-fact rows are kept as reference upper bounds for what perfect-recall retrieval could achieve. The gap between UaC and live baselines widens on hard arithmetic; on hard scenarios, UaC's CI lower bound (64.0\%) clears the upper bound of A-MEM and UaC-no-alerts, but overlaps with live Mem0 and simulated Mem0.}
\label{tab:active-hard}
\small
\begin{tabular}{@{}lcc@{}}
\toprule
\textbf{System} & \textbf{Standard (CI)} & \textbf{Hard (CI)} \\
\midrule
\textbf{UaC + pipeline} & \textbf{100\%} [91.2, 100] & \textbf{85.0\%} [64.0, 94.8] \\
Mem0 (live, mem0ai 1.0.5) & 90.0\% [76.9, 96.0] & 80.0\% [58.4, 91.9] \\
\quad Mem0 (simulated, ref.) & 92.5\% [80.1, 97.4] & 65.0\% [43.3, 81.9] \\
Full Context (live) & --- & 55.0\% [34.2, 74.2] \\
UaC (no alerts) & 52.5\% [37.5, 67.1] & 45.0\% [25.8, 65.8] \\
A-MEM (live) & --- & 30.0\% [14.5, 51.9] \\
\quad A-MEM (simulated, ref.) & 85.0\% [70.9, 92.9] & --- \\
\bottomrule
\end{tabular}
\end{table}

On hard scenarios (Table~\ref{tab:active-hard}) the gap widens for most baselines: simulated Mem0 drops 27.5pp, live A-MEM falls to 30\%, and Full~Context reaches only 55\%---even with all raw text, the thinking model struggles with multi-step arithmetic. UaC drops 15pp to 85\%, and its CI lower bound (64.0\%) clears A-MEM, Full~Context, and no-alerts UaC. The one comparison we cannot resolve is live Mem0 (80\%): CIs overlap heavily ([58.4, 91.9] vs.\ [64.0, 94.8]) and McNemar gives $p{=}0.69$ (3 UaC-only correct vs.\ 2 Mem0-only out of 20). We therefore explicitly \emph{do not} claim a significant gap over live Mem0 on the hard set---reaching $\alpha{=}0.05$ at this 5pp effect would need $n \approx 200$, an order of magnitude more than we have authored (flagged in Section~\ref{sec:limits}). The pipeline separates UaC from four of five baselines at $p<0.01$; the fifth remains inconclusive at $n{=}20$.

\subsection{Ablation Study}
\label{sec:ablation}

\begin{table}[htbp]
\centering
\caption{Ablation on the first 5-conversation LOCOMO subset (300 QAs) and 40 Active Service scenarios. Each row names a memory configuration rather than a version number. The study was run on the 5-conv subset before the headline was extended to 10 conversations; the \emph{Two-phase} row's 78.0\% here corresponds to the same architecture whose 10-conversation score is 78.8\% in Table~\ref{tab:locomo}.}
\label{tab:ablation}
\begin{tabular}{@{}lccp{5cm}@{}}
\toprule
\textbf{Configuration} & \textbf{LOCOMO} & \textbf{Active Service} & \textbf{Key change} \\
\midrule
Basic 3-tier & 56.7\% & 30.0\% & Code + basic archive RAG \\
Flat facts & 75.7\% & 37.5\% & Append-only fact extraction (+19.0pp recall) \\
Incremental code & 65.7\% & 40.0\% & Single code file, overwritten each session \\
\textbf{Two-phase (full UaC)} & \textbf{78.0\%} & 67.5\% & Append-only facts + periodic restructuring \\
\quad + constraint pipeline & 78.0\% & \textbf{100\%} & Adds the generate--verify--review loop \\
\bottomrule
\end{tabular}
\end{table}

\begin{figure}[htbp]
\centering
\includegraphics[width=0.9\textwidth]{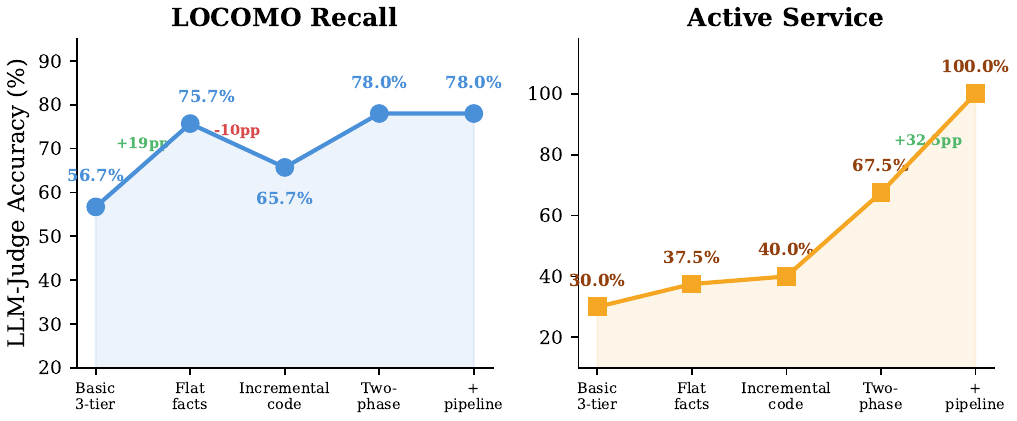}
\caption{Ablation study showing the impact of each architectural decision (5-conv LOCOMO, n=300). \textbf{Left:} LOCOMO recall improves +19pp when adding append-only extraction (Basic 3-tier $\to$ Flat facts), drops $-$10pp under incremental code overwrite (Flat facts $\to$ Incremental code), then recovers and exceeds with two-phase separation. \textbf{Right:} Active Service improves steadily with code structure, with the constraint pipeline adding +32.5pp on top of the two-phase architecture.}
\label{fig:ablation}
\end{figure}

The ablation (Table~\ref{tab:ablation}, Figure~\ref{fig:ablation}) reveals four key findings:

\textbf{Finding 1: Append-only extraction is the single biggest improvement.} Adding append-only fact extraction (Basic 3-tier $\to$ Flat facts) gains +19.0pp on LOCOMO. Never overwriting facts prevents information loss during incremental state updates.

\textbf{Finding 2: Code structure alone hurts recall if done incrementally.} Switching from flat facts to incremental code (Flat facts $\to$ Incremental code) loses 10.0pp on LOCOMO. Each session's code rewrite drops earlier facts. The \texttt{notes} field (intended as a safety net) becomes an escape hatch---the LLM dumps facts as strings instead of typing them.

\textbf{Finding 3: Two-phase separation solves the overwrite problem.} The two-phase design combines the append-only coverage of \emph{Flat facts} with the typed structure of \emph{Incremental code}, gaining +12.3pp over Incremental code and +2.3pp over Flat facts. The structuring step generates code from the \emph{complete} fact corpus, avoiding incremental information loss.

\textbf{Finding 4: The constraint pipeline is orthogonal to LOCOMO QA.} Adding the constraint-generation pipeline on top of the two-phase architecture leaves LOCOMO QA accuracy unchanged at 78.0\% but lifts Active Service alert detection from 67.5\% to 100\%. The pipeline is a separate read-time mechanism that operates over the same structured state; it does not affect retrieval quality.

\subsection{Retrieval-Channel Ablation}
\label{sec:channel-ablation}

UaC's retrieval combines three channels---structured code state, fact-vector top-20, raw archive top-10 (Section~\ref{sec:retrieval}). A reasonable concern with the headline LOCOMO number is that the gain over baselines could come from ``three retrieval channels'' rather than from the structured representation itself. To isolate each channel's marginal contribution, we run UaC on the same 5-conversation, 300-QA LOCOMO subset under three leave-one-out configurations: $-$STATE (only FACTS+ARCHIVE), $-$FACTS (only STATE+ARCHIVE), and $-$ARCHIVE (only STATE+FACTS). Each conversation is ingested once and the structured state cached, so Phase~1 and Phase~2 costs are constant across configurations.

\begin{table}[!ht]
\centering
\caption{Retrieval-channel leave-one-out ablation on the first-5-conversation LOCOMO subset (300 QAs). \textbf{Full} reproduces the 5-conv UaC number (78.0\%); each subsequent column reports UaC with one of the three retrieval channels removed at answer time. McNemar's exact test on the per-question correctness vector compares Full vs.\ each lesion. The ablation was conducted at 5-conv scale; on the full 10-conv benchmark the corresponding Full number is 78.8\% (Table~\ref{tab:locomo}).}
\label{tab:channel-ablation}
\small
\begin{tabular}{@{}lrrrr@{}}
\toprule
\textbf{Conversation} & \textbf{Full} & $-$\textbf{STATE} & $-$\textbf{FACTS} & $-$\textbf{ARCHIVE} \\
\midrule
conv-26 & 75.0\% & 83.3\% & 70.0\% & 75.0\% \\
conv-30 & 90.0\% & 85.0\% & 86.7\% & 80.0\% \\
conv-41 & 80.0\% & 80.0\% & 73.3\% & 71.7\% \\
conv-42 & 76.7\% & 68.3\% & 48.3\% & 61.7\% \\
conv-43 & 68.3\% & 66.7\% & 65.0\% & 65.0\% \\
\midrule
\textbf{Overall} & \textbf{78.0\%} & 76.7\% & 68.7\% & 70.7\% \\
$\Delta$ vs.\ Full & --- & $-1.3$pp & $-9.3$pp & $-7.3$pp \\
McNemar $p$ vs.\ Full & --- & $0.67$ & $0.0008$ & $0.008$ \\
\bottomrule
\end{tabular}
\end{table}

The ablation (Table~\ref{tab:channel-ablation}) splits the three channels into two qualitative groups.

\textbf{Fact-vector and raw-archive channels are both significant individually.} Removing the fact-vector channel drops accuracy 9.3pp (McNemar $p{=}0.0008$); removing the raw archive drops 7.3pp ($p{=}0.008$). Both are clearly load-bearing for LOCOMO-style recall: fact vectors carry the long tail of details that did not make it into the typed schema, while the archive preserves verbatim phrasing for direct-quote questions.

\textbf{The structured code state is not statistically significant on LOCOMO recall} ($-1.3$pp, McNemar $p{=}0.67$). This is the part of the finding we want to be honest about: on pure recall benchmarks where the answer can almost always be found by retrieval over facts plus raw text, the typed Python state is roughly redundant with the fact channel. The per-conversation breakdown is informative---STATE \emph{helps} on conv-42 ($-8.3$pp when removed) and \emph{hurts} on conv-26 ($+8.3$pp when removed)---which is consistent with the structured state introducing compression that occasionally drops a detail the question needed, then routing the LLM toward typed objects rather than raw phrasing.

\textbf{Where the structured state pays off is elsewhere in the paper.} The headline analytical-inference numbers (Section~\ref{sec:analytical}) are unreachable without a code-readable representation: retrieval-based systems collapse to 6--43\% on the same questions UaC answers at 99\%. The constraint pipeline in Active Service (Section~\ref{sec:active-service}) cannot run without it either. So the right way to read Table~\ref{tab:channel-ablation} is not as evidence that the structured state is useless, but as a quantification of where it does and does not move the needle: it is roughly neutral for LOCOMO-style recall and decisive for the two capability tiers retrieval cannot serve. The fact-vector and archive channels remain the right tools for surface-level recall, and the three-channel design composes those two retrieval modes with the structure-enabled analytical and constraint modes.

\subsection{Modularity and Progressive Disclosure}
\label{sec:modularity}

We finally evaluate two design principles introduced in Section~\ref{sec:method} that the standard benchmarks do not isolate: \emph{Modularity by Life Domain} and \emph{Progressive Disclosure}. We synthesize a multi-domain user state with 10 domains (trips, contacts, meals, transactions, sleep, workouts, books, medical visits, meetings, purchases) at 50 records each (500 total), then ask 100 single-domain questions (10 per domain---the full question-pattern coverage from the analytical benchmark). Three loading strategies share the same Python REPL backend and Gemini~3 Flash model:

\textbf{Monolithic:} the full 500-record state is inlined into the LLM system prompt every query (the naive ``always-dump'' baseline).

\textbf{Modular:} only a list of available domain names is in the system prompt. The LLM calls a \texttt{load\_domain(name)} REPL helper to materialize the specific domain it needs.

\textbf{Manifest+routing:} the system prompt additionally contains a one-line summary per domain ($\sim$50 tokens each); the LLM uses the manifest to route then calls \texttt{load\_domain}.

\begin{table}[!ht]
\centering
\caption{Modularity / Progressive Disclosure ablation: 100 single-domain questions over a 10-domain, 500-record user state. Monolithic dumps the full state in every prompt; Modular and Manifest+routing load only the relevant domain on demand.}
\label{tab:modularity}
\small
\begin{tabular}{@{}lrrrr@{}}
\toprule
\textbf{Strategy} & \textbf{Acc} & \textbf{Prompt tokens} & \textbf{Total cost} & \textbf{Per case} \\
\midrule
Monolithic (always dump) & 97.0\% & 10,155,765 & \$3.64 & 36.4 m\$ \\
Modular (load on demand) & \textbf{98.0\%} & 426,856 & \$0.25 & 2.5 m\$ \\
Manifest+routing & 87.0\% & 624,922 & \$0.38 & 3.8 m\$ \\
\bottomrule
\end{tabular}
\end{table}

\begin{figure}[htbp]
\centering
\includegraphics[width=0.85\textwidth]{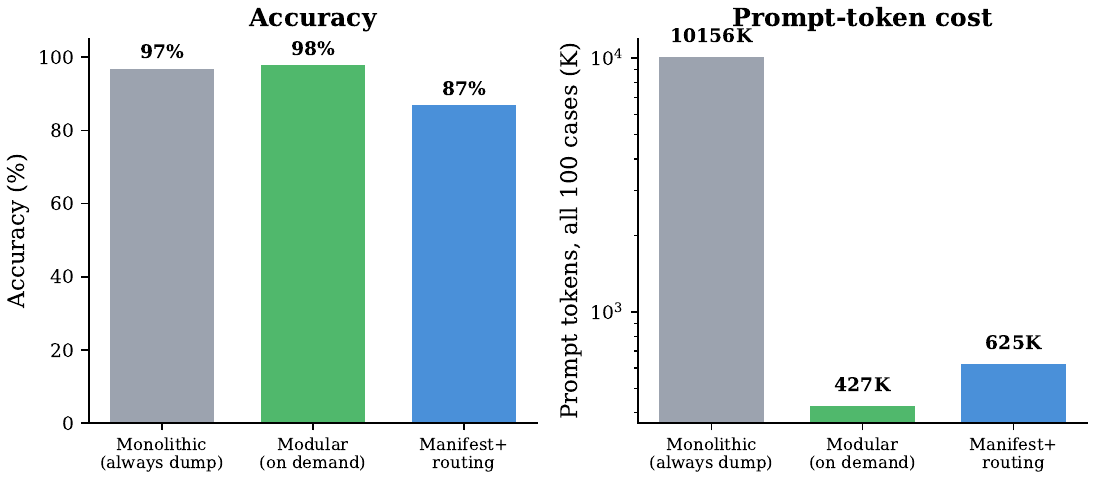}
\caption{Modularity / Progressive Disclosure ablation (100 cases, 500 records). \textbf{Left:} modular loading matches or beats the monolithic upper bound; the manifest variant trails by 10pp due to mis-routing. \textbf{Right:} prompt-token count drops 24$\times$ when only the relevant domain is loaded.}
\label{fig:modularity}
\end{figure}

\textbf{Modular loading matches monolithic on accuracy} (Table~\ref{tab:modularity}, Figure~\ref{fig:modularity}; 98.0\% vs.\ 97.0\%, one-question difference well inside the Wilson interval) at \textbf{14.9$\times$ lower dollar cost} (\$0.25 vs.\ \$3.64 over 100 cases; the underlying prompt-token count drops 24$\times$). At 500 records the monolithic prompt exceeds the LLM's effective working window and occasionally retrieves the wrong record from its own context; loading only the query-relevant domain shrinks the prompt to $\sim$50 records and the model reads it cleanly. Manifest+routing loses 10pp because the LLM occasionally mis-routes from the one-line summaries. Monolithic cost grows linearly with state size; modular cost is bounded by the largest single domain---a structural advantage as user state accumulates. This ablation also reverses the earlier 30-case finding (where monolithic was perfect and modular trailed by 6.7pp): below the working-memory ceiling, modularity's accuracy cost vanishes as state scales, and it becomes \emph{Pareto-preferable} at 500 records.

\section{Discussion}

\textbf{SOTA retrieval mechanisms are additive, not competitive.} A UaC + MemMachine-style episode-retrieval hybrid (Appendix~\ref{app:additivity}) reaches 81.7\% on the 5-conversation LOCOMO subset vs.\ 78.0\% plain UaC (+3.7pp; McNemar $p{=}0.14$, CIs overlap). The gain concentrates on the two hardest conversations where UaC's structured state had compressed away the relevant detail. We report this as a directional signal rather than a guaranteed effect at $n{=}300$, but the implication is that episode retrieval from MemMachine~\citep{wang2026memmachine} or EverMemOS~\citep{hu2026evermemos} can be layered on top of UaC rather than swapped for it.

\textbf{Domain knowledge is LLM-generated; infrastructure is human-designed.} Schemas, constraints, and domain partitioning are all produced by the LLM and evolve with the user. The engineering scaffolding---two-phase pipeline, generate--verify--review loop, sandbox boundary---is human-designed and fixed. The interpreter is a \emph{tool the LLM uses to verify its own outputs}, not a rule layer it must obey, which keeps the system open-ended as user state grows.

\textbf{No human-designed schema: aligning with the scaling law.} A defining property of UaC is that the data model itself is \emph{not} hand-authored. Relational and document stores require a human to commit to a schema before any data arrives; that schema then bounds what the system can represent and is costly to evolve. UaC inverts this: the dataclasses, domain partitioning, and constraints are written autonomously by the LLM at structuring time (Section~\ref{sec:method}) and regenerated from the full fact corpus as the user's life changes, so the representation is flexible and self-evolving rather than fixed. We read this as a deliberate bet on Sutton's ``bitter lesson''~\citep{sutton2019bitter}---that hand-engineered structure tends, in the long run, to be overtaken by general methods that scale with data and compute. Instead of encoding a fixed ontology, UaC leans on the single capability current LLMs are strongest at---code generation---and lets the model design the data structures that fit each user. The absence of a human-designed schema is therefore a feature, not a pitfall: the only human contribution is the scaffolding, not the structure of the memory itself.

\textbf{Memory safety and RL-trainability.} PersistBench~\citep{pulipaka2026persistbench} reports 53\% cross-domain leakage and 97\% sycophancy in persistent memory; UaC's domain separation limits leakage, version control provides audit trails, and explicit code enables selective forgetting. Because memory operations are file-system actions, UaC is also directly compatible with RL-based memory training~\citep{yan2025memoryr1, yu2026agemem, wang2025memalpha}.

\textbf{Limitations.}\label{sec:limits}
\begin{itemize}[leftmargin=*]
\item \emph{Write-time overhead.} UaC runs two LLM passes (Phase~1 + Phase~2) where flat-fact systems run one. Structuring costs $\sim$7.6--101.8~m\$/case at $N{\in}\{20,500\}$ (Table~\ref{tab:analytical-cost-amortized}). The write-path ratio is workload-dependent; we report only per-case costs and the amortized $15\times$ read-time advantage at $K{=}100$. Phase~2's all-at-once regeneration also implies a scalability ceiling at hundreds of sessions, which is why we flag hierarchical structuring as future work; the Modularity ablation (Section~\ref{sec:modularity}) is a partial step.
\item \emph{Structured state is roughly neutral for pure recall.} The channel ablation (Section~\ref{sec:channel-ablation}) shows the typed Python state contributes only $-1.3$pp on LOCOMO QA ($p{=}0.67$). Its value is concentrated in analytical inference and the constraint pipeline---a stated contribution (Section~\ref{sec:intro}), not just a caveat. A reader who only needs single-hop recall does not need UaC's structuring step. Likewise the residual 1.0pp gap to Full~Context on LOCOMO ($p{=}0.65$) is dominated by single-hop minor details (specific book titles, exact early-session dates) that extraction occasionally compresses away. A buggy constraint will also fire deterministically every time---correct computation is not correct logic.
\item \emph{Active Service depends on the pipeline.} Without pre-computed alerts UaC drops to 52.5\%---the pipeline, not the representation, is the decisive factor on this benchmark. The 20-scenario hard set is also underpowered against the strongest baseline (live Mem0 CIs overlap UaC's, McNemar $p{=}0.69$); $n \approx 200$ would be needed to resolve the 5pp effect at $\alpha{=}0.05$.
\item \emph{Analytical benchmark is synthetic and single-query.} Records are schema-clean (real-world data will be noisier, which we expect to favor UaC over FC+REPL since the baseline's parsing fragility scales with noise); the single-query design is also the worst case for UaC (Section~\ref{sec:analytical} shows structuring pays back after $\sim$3 queries).
\item \emph{Reimplementation fidelity and infrastructure confound.} Our same-backbone reproductions replace published retrieval stacks (e.g., Mongo+Elasticsearch+Milvus; Tempr/pgvector) with a single ChromaDB collection (Appendix~\ref{app:reimpl}). The published numbers under stronger backbones (GPT-4.1-mini, Gemini-3 Pro) are the architecture ceiling, not direct competitors; we cannot fully decompose how much of our same-backbone gap comes from representation vs.\ retrieval stack. Reproducing each SOTA with native infrastructure is a follow-up.
\item \emph{Cross-LLM coverage and judge bias.} Portability is verified on Gemini~3 Flash and GPT-5.4 only; smaller (7--8B) code generators may produce lower-quality structured code. The Gemini answer LLM is also the same-family judge; we partially address self-bias with a Claude rejudging of all 7{,}700 predictions (Section~\ref{sec:judge-check}, $\kappa \geq 0.74$ on every system).
\item \emph{Sandboxing required.} Storing user state as executable code requires sandboxed execution and strict isolation to prevent cross-user leakage or code injection.
\end{itemize}

\section{Conclusion}

User as Code reframes user memory as a modular software project rather than a passive database. The architectural insight the ablation isolates is that \emph{memorizing and structuring must be separate concerns}: append-only extraction preserves coverage (+19pp on LOCOMO over a code-only baseline), and periodic structuring adds the typed surface a Python interpreter can read (a further +12.3pp over incremental code rewrites). Once memory is code, basic recall, analytical inference, and proactive constraint checking become three uses of the same medium---which is why the standard-benchmark, analytical, and Active Service results move together.

\textbf{Future work.} Three directions follow from the limitations: (i)~hierarchical or incremental structuring that scales past Phase~2's all-at-once regeneration; (ii)~RL-trained constraint generation that learns which ad-hoc checks to promote to persistent; (iii)~integration with SOTA episode-retrieval mechanisms whose additivity we already establish (Appendix~\ref{app:additivity}).

\textbf{A closing thought.} The Chinese word for memory is written with two characters---\emph{j\`i} (to record, the encoding of information from its source) and \emph{y\`i} (to recall, the retrieval of it). The two are not independent: what can be recalled is bounded by how it was recorded. We read User as Code as an argument that the representation sitting between them is the real lever---a record-side structure expressive enough that the recall side can read back what it needs with both high recall and high precision. Typed, executable code is our bet on what that structure should be: a medium in which memorizing and recalling meet, and one that an interpreter, not just a similarity search, can run.

\section*{Acknowledgements}

Pine Copilot, Claude Code, and Claude Opus~4.8 were used during this research.

\appendix

\section{Reimplementation Details for SOTA Baselines}
\label{app:reimpl}

This appendix documents what our reimplementations of MemMachine, EverMemOS, and Hindsight match faithfully vs.\ what they approximate, so the same-backbone numbers in Tables~\ref{tab:locomo}--\ref{tab:longmemeval} can be read with the right level of skepticism.

\subsection{Shared infrastructure}

All three baselines share UaC's evaluation harness: Gemini~3 Flash (\texttt{gemini-3-flash-\allowbreak preview}) for every LLM call, ChromaDB with the default embedding for vector retrieval, the same generous-scoring LLM-as-Judge (Section~\ref{sec:experiments}), and the same per-question reset. The published systems use heavier infrastructure---MongoDB$+$Elasticsearch$+$Milvus (MemMachine), Tempr/pgvector (Hindsight), or unspecified hybrid stores---which we deliberately replace with a minimal common stack to isolate the representation as the variable. Per-baseline source files: \texttt{run\_locomo\_memmachine.py}, \texttt{evermemos\_lite.py}, \texttt{hindsight\_lite.py}.

\subsection{MemMachine (\texttt{run\_locomo\_memmachine.py}, \texorpdfstring{$\sim$245}{\textasciitilde 245} LOC)}

\textbf{Faithful to the description.} (i)~Episode-level storage: every conversation turn is indexed verbatim as a single sentence \texttt{[session\_id date] speaker: text}, and full-ordered episodes are kept for context expansion. (ii)~Nucleus retrieval: top-$k$ dense match with $k{=}30$. (iii)~Contextual expansion: $\pm 3$ surrounding sentences around each nucleus hit, deduplicated and re-ordered by original position. (iv)~The retrieved expanded context is passed verbatim to Gemini~3 Flash for answering.

\textbf{Approximated.} The MemMachine paper specifies neither the exact embedding model nor the rerank stage; we use ChromaDB's default embedding (\texttt{all-MiniLM-L6-v2}-equivalent) and skip cross-encoder reranking. The published system also pairs sentence-level with profile-level memory; we omit profile memory because the LOCOMO conversations are short enough that per-question retrieval already accesses the relevant content. Both omissions plausibly reduce our MemMachine score relative to the original paper.

\subsection{EverMemOS (\texttt{evermemos\_lite.py}, \texorpdfstring{$\sim$394}{\textasciitilde 394} LOC)}

\textbf{Faithful to the description.} The three-phase pipeline from the paper abstract: (i)~Phase~1 ``Episodic Trace Formation''---each session is segmented into MemCells, with an LLM extracting atomic facts and Foresight signals (predicted future information needs). (ii)~Phase~2 ``Semantic Consolidation''---MemCells are clustered into thematic MemScenes via similarity; a scene-level summary is produced and a User Profile of stable attributes is updated. (iii)~Phase~3 ``Reconstructive Recollection''---query embedding $\to$ MemScene-level filter (\texttt{SCENE\_TOP\_K=3}) $\to$ MemCell-level dense retrieval (\texttt{CELL\_TOP\_K=20}) $\to$ BM25 keyword channel (\texttt{BM25\_TOP\_K=20}) $\to$ rerank by recency and Foresight overlap (\texttt{FORESIGHT\_BOOST=0.5}) $\to$ atomic-fact aggregation $\to$ context assembly.

\textbf{Approximated.} The paper specifies neither the exact embedding model nor the clustering algorithm parameters; we use ChromaDB defaults and a similarity-threshold linkage clustering. Foresight signal extraction is implemented but the trigger threshold is a default we picked rather than one from the paper. We do not implement the long-horizon engram-decay schedule (the paper describes it as ``cell-level forgetting''); on a 5-conversation benchmark this affects nothing materially, but on month-long deployments it would.

\subsection{Hindsight (\texttt{hindsight\_lite.py}, \texorpdfstring{$\sim$420}{\textasciitilde 420} LOC)}

\textbf{Faithful to the description.} (i)~The 10-tuple Fact node \texttt{(subject, body, time, embedding, $\tau_s$, $\tau_e$, $\tau_m$, label, confidence, extras)} is built explicitly. (ii)~The four logical fact networks (World, Experience, Opinion, Observation) are constructed via labeled extraction. (iii)~Coarse-grained chunking of conversation turns into 2--5 facts per turn via an LLM extractor. (iv)~Four-channel retrieval: semantic (top 30), BM25 (top 30), graph spreading activation (2 hops, decay 0.6), and temporal proximity (top 30), all fused with Reciprocal Rank Fusion ($k{=}60$). (v)~Cross-encoder rerank with \texttt{cross-encoder/ms-marco-MiniLM-L-6-v2}, final 20 facts, packed within a 4000-token budget.

\textbf{Approximated / omitted.} (i)~The CARA disposition layer (controllable agent response affect) is omitted---it shapes the agent's voice and beliefs, not its retrieval. (ii)~Opinion formation/reinforcement (the paper's belief-update dynamics) is omitted; we keep facts as immutable records with a confidence field. (iii)~The published Tempr / pgvector store is replaced with ChromaDB plus an in-memory BM25 index. The paper's reported 91.4\% on LongMemEval uses a Gemini-3 \emph{Pro} backbone in the full pipeline; our 73.0\% on the full 500-question LongMemEval under Gemini-3 \emph{Flash} with the minimal infrastructure isolates the representation effect from those amplifying factors.

\subsection{Backbone caveat}

Gemini~3 \emph{Flash} is the lower-cost sibling of Gemini~3 Pro and is weaker on long-context reasoning. The published EverMemOS, Hindsight, and MemMachine numbers use stronger backbones (GPT-4.1-mini, Gemini-3 Pro, mixed proprietary) and are not directly comparable to ours. UaC runs under the same handicaps as the baselines here, so the \emph{relative} ranking is interpretable; the \emph{absolute} numbers are floor estimates of what each architecture can do.

\subsection{UaC hyperparameters}

For completeness, the UaC hyperparameters used throughout the paper are: extraction LLM Gemini~3 Flash, thinking budget 8192; structuring LLM Gemini~3 Flash, thinking budget 16{,}384, no \texttt{max\_output\_tokens} cap; answer-time LLM Gemini~3 Flash, thinking budget 2048, temperature 1.0; vector store ChromaDB with cosine similarity and the default embedding; retrieval combines (a)~the structured Python state truncated to 6{,}000 chars if longer, (b)~top-20 facts from the per-fact vector index, (c)~top-10 raw conversation excerpts from the archive index. Judge LLM Gemini~3 Flash, thinking budget 256.

\section{Additivity of SOTA Retrieval Mechanisms}
\label{app:additivity}

Section~\ref{sec:experiments} compares UaC against MemMachine, EverMemOS, and Hindsight as standalone systems; the comparison is between representations under one backbone. A separate question is \emph{additivity}: if we keep UaC's structured-Python representation and \emph{also} add a SOTA retrieval mechanism on top, does the score improve?

\textbf{Setup.} We construct a UaC + MemMachine variant (\texttt{user\_as\_code\_plus\_mm.py}) that retains UaC's three-channel retrieval (structured code, fact-vector, raw archive) and additionally appends a MemMachine-style episode context (sentence-level dense retrieval with $\pm 3$ contextual expansion, top-30 nucleus) as a fourth channel. Everything else is identical: same conversations (5-conversation LOCOMO subset, 300 QAs), same Gemini~3 Flash backbone, same judge, same prompts.

\textbf{Result.} The hybrid reaches \textbf{81.7\%} on the 5-conversation LOCOMO subset, vs.\ 78.0\% for plain UaC---a +3.7pp gain. McNemar's exact test on the per-question correctness gives $p{=}0.14$ (29 hybrid-only correct vs.\ 18 UaC-only correct out of 300), and Wilson 95\% CIs $[76.9, 85.7]$ vs.\ $[73.0, 82.3]$ overlap, so we treat this as a clear directional signal rather than a guaranteed effect at $n{=}300$.

\textbf{Interpretation.} The two mechanisms are partially complementary: structured Python state surfaces typed facts (dates, names, numbers), while episode retrieval surfaces the surrounding conversational context. The gain is concentrated in the two harder conversations (conv-41 +8.3pp, conv-43 +8.3pp) where plain UaC was weakest, suggesting that episode context most helps when the structured state has compressed away the relevant detail. Per-conversation results: conv-26 75.0\%/75.0\% (tie), conv-30 90.0\%/93.3\% (+3.3pp), conv-41 80.0\%/88.3\% (+8.3pp), conv-42 76.7\%/75.0\% ($-1.7$pp), conv-43 68.3\%/76.7\% (+8.3pp). The cost is roughly doubled per query because the MM index is added on top of UaC's three existing channels; in practice an implementation should switch the MM channel on only when the structured state's similarity to the query is below threshold.

\section{Cross-Family Judge Rejudging}
\label{sec:judge-check}

The headline LOCOMO and LongMemEval numbers in Section~\ref{sec:experiments} use a Gemini~3 Flash LLM-as-Judge, which is also the answer-generation backbone for every system. Self-bias in same-family LLM-as-Judge setups is a documented concern; we therefore re-judge \emph{every} prediction from \emph{every} system under Claude Opus~4.7 via OpenRouter, with the same generous-scoring prompt. The rejudging covers all 600 LOCOMO QAs and all 500 LongMemEval QAs per system, for the full 7 systems and a total of $n{=}7{,}700$ Claude calls.

\begin{table}[!ht]
\centering
\caption{Cross-family judge rejudging on the full LOCOMO (n=600/sys) and LongMemEval (n=500/sys) benchmarks. ``Gemini'' is the same-family judge used in the headline numbers; ``Claude'' is the cross-family judge. Agreement is row-level fraction; Cohen's $\kappa$ is the chance-corrected agreement (Landis--Koch: $\kappa{\in}[0.61,0.80]$ substantial, $[0.81,1.0]$ almost perfect).}
\label{tab:judge-rejudge}
\small
\begin{tabular}{@{}llrrrrr@{}}
\toprule
\textbf{Dataset} & \textbf{System} & \textbf{Gemini} & \textbf{Claude} & $\Delta$ & \textbf{Agree} & $\kappa$ \\
\midrule
LOCOMO & UaC       & 78.8\% & 75.7\% & $-3.1$ & 91.8\% & 0.77 \\
LOCOMO & Full Context & 79.8\% & 77.5\% & $-2.3$ & 93.0\% & 0.79 \\
LOCOMO & MemMachine   & 72.7\% & 68.2\% & $-4.5$ & 91.8\% & 0.80 \\
LOCOMO & Hindsight    & 69.7\% & 65.3\% & $-4.4$ & 90.0\% & 0.77 \\
LOCOMO & EverMemOS    & 55.5\% & 50.2\% & $-5.3$ & 91.3\% & 0.83 \\
LOCOMO & A-MEM        & 51.8\% & 47.0\% & $-4.8$ & 92.2\% & 0.84 \\
LOCOMO & Mem0         & 29.3\% & 23.3\% & $-6.0$ & 90.0\% & 0.74 \\
\midrule
LME-500 & UaC       & 83.0\% & 83.0\% & $\pm 0.0$ & 98.0\% & 0.93 \\
LME-500 & Full Context & 85.4\% & 85.0\% & $-0.4$ & 97.6\% & 0.91 \\
LME-500 & MemMachine   & 84.8\% & 84.0\% & $-0.8$ & 99.2\% & 0.97 \\
LME-500 & EverMemOS    & 76.4\% & 74.8\% & $-1.6$ & 97.6\% & 0.94 \\
LME-500 & Hindsight    & 73.0\% & 72.6\% & $-0.4$ & 95.6\% & 0.89 \\
LME-500 & A-MEM        & 49.6\% & 48.0\% & $-1.6$ & 97.6\% & 0.95 \\
LME-500 & Mem0         & 23.8\% & 23.2\% & $-0.6$ & 97.8\% & 0.94 \\
\bottomrule
\end{tabular}
\end{table}

Three things from Table~\ref{tab:judge-rejudge}:

\textbf{(1)~Substantial-to-almost-perfect agreement across all 7 systems.} On LOCOMO, every $\kappa$ is in the substantial-to-almost-perfect range ($0.74 \leq \kappa \leq 0.84$). On LongMemEval, every $\kappa$ is almost-perfect ($0.89 \leq \kappa \leq 0.97$), meaning the cross-family judge agrees with the same-family judge on essentially every prediction. The rankings under the two judges are identical on both benchmarks.

\textbf{(2)~Claude is uniformly slightly stricter, but the relative ranking is preserved and UaC's gaps over the lower systems widen.} On LOCOMO, Claude scores every system 2--6pp lower than Gemini; on LongMemEval, the difference is under 2pp. Importantly, the UaC--MemMachine gap on LOCOMO grows from $+6.1$pp under Gemini to $+7.5$pp under Claude, and the UaC--Mem0 gap grows from $+49.5$pp to $+52.4$pp. So the same-family judge is, if anything, slightly favorable to the lower-scoring competitors; the cross-family judge sharpens UaC's lead rather than erasing it.

\textbf{(3)~Cross-judge drops correlate with prediction ambiguity, not with system identity.} On LOCOMO, Mem0 has the largest drop ($-6.0$pp), MemMachine the second-largest, and the typed-state UaC sits in the middle. On LongMemEval, where answers are more often verbatim spans pulled from the haystack, all drops are under 2pp and the agreement is uniformly almost-perfect. The pattern matches the intuition that predictions phrased as terse fragments (``yesterday'', short noun phrases) are more sensitive to judge generosity, while typed or fully-formed answers are less so.

The headline rankings on both LOCOMO and LongMemEval are stable under the cross-family judge---which is the load-bearing claim of this appendix. We treat this as evidence that the same-family judge does not materially distort the comparison.

\section{Phase-2 Failure-Mode Analysis}
\label{sec:phase2-failures}

Phase~2 is an LLM call: it can drop facts, produce non-parseable code, or store dates as strings instead of typed \texttt{date()} objects. We audit each of these failure modes by running Phase~2 on all five LOCOMO conversations and inspecting the generated typed Python. The four checks are: \emph{(i)~parse}---does the output parse as valid Python; \emph{(ii)~date typing}---are dates stored as \texttt{date(\dots)} rather than strings; \emph{(iii)~notes-bucket usage}---do facts that do not fit typed fields land in a \texttt{notes: list[str]} so they are still in the file; \emph{(iv)~dropped facts}---are there input facts whose distinctive content tokens (rare nouns, numbers, year-formatted dates) appear nowhere in the generated code, even in notes.

\begin{table}[!ht]
\centering
\caption{Phase-2 failure-mode audit on the 5 LOCOMO conversations used in the headline numbers. Phase~1 produces $\sim$1.5K--2.6K facts per conversation; Phase~2 regenerates typed Python from the complete fact corpus. Drop rate is computed against the count of facts with at least one distinctive content token (i.e., facts that are not pure stopwords).}
\label{tab:phase2-failures}
\small
\begin{tabular}{@{}lrrrrrrr@{}}
\toprule
\textbf{Conv} & \textbf{Facts} & \textbf{Parse} & \textbf{Dataclass} & \textbf{Notes} & \textbf{Dates} & \textbf{Dropped} & \textbf{Drop \%} \\
 & & & \textbf{instances} & \textbf{entries} & \textbf{typed/str} & & \\
\midrule
conv-26 & 1,631 & \checkmark & 2 & 49 & 9 / 0 & 0 & 0.0\% \\
conv-30 & 1,433 & \checkmark & 4 & 133 & 11 / 0 & 0 & 0.0\% \\
conv-41 & 2,483 & \checkmark & 9 & 146 & 11 / 0 & 0 & 0.0\% \\
conv-42 & 2,173 & \checkmark & 11 & 44 & 8 / 0 & 0 & 0.0\% \\
conv-43 & 2,553 & \checkmark & 11 & 51 & 5 / 0 & 18 & 0.7\% \\
\midrule
\textbf{Total} & \textbf{10,273} & \textbf{5/5} & \textbf{37} & \textbf{423} & \textbf{44 / 0} & \textbf{18} & \textbf{0.18\%} \\
\bottomrule
\end{tabular}
\end{table}

\textbf{Parseability.} Every generated file parses as valid Python (5/5). Phase~2 has not produced syntax errors on any LOCOMO conversation in our audit. This matters because the answer-time channel passes the structured state to the LLM verbatim; an unparseable file would silently corrupt the structured-state retrieval.

\textbf{Date typing.} Across all 5 conversations, the generated code contains 44 \texttt{date(\dots)} expressions and zero string-formatted dates. The 16{,}384-token thinking budget appears sufficient for the LLM to follow the explicit date-typing instruction reliably. This matters specifically for LongMemEval's temporal-reasoning category, which exercises date arithmetic.

\textbf{Dropped facts.} Across the 10{,}273 distinctive-token facts audited in Table~\ref{tab:phase2-failures}, 18 are dropped (overall rate 0.18\%). All 18 are from conv-43, which has the largest fact corpus (2{,}553 facts) and produced the \emph{smallest} structured code (6{,}329 chars, vs.\ 12{,}222 for conv-30 with fewer facts). The pattern is consistent: when the structuring LLM is overloaded, it compresses by abstracting away atomic details rather than by losing structure or types. The 18 dropped facts are all from the same session (session\_27) and concern minor personal-state details (e.g., ``Tim finds learning German to be tough''). The notes bucket caught 423 facts that did not fit typed fields across the corpus---the safety net is doing its job.

\textbf{Implications.} Phase~2 has three failure axes: parse, type, content. The first two are clean on this benchmark (0/5 and 0/44 respectively). Content-loss is non-zero but small (0.18\% overall on the audited five conversations) and concentrated on the longest input. This bounds the headline LOCOMO gap to Full Context: of the 1.0pp UaC-vs-Full Context gap on the full 10-conversation benchmark (Section~\ref{sec:experiments}), at most $\sim$0.2pp is explained by Phase~2 dropping facts the question needed; the remainder is in the retrieval layer (channel ablation in Section~\ref{sec:channel-ablation}). At even larger scales the truncation behavior documented in Section~\ref{sec:phase2-scalability} would dominate over the LLM-level compression we see here, which is why we flag hierarchical structuring as the right scaling strategy.

\section{Phase-2 Structuring Cost at Scale}
\label{sec:phase2-scalability}

Phase~2 is a single LLM call that regenerates structured Python from the entire fact corpus accumulated so far. To measure how its cost behaves outside the 19-session LOCOMO range, we run Phase~2 on synthetically lengthened corpora at $\{19, 50, 100, 200\}$ sessions, seeded from the actual conv-26 Phase~1 fact list (1{,}569 facts) and extended by appending shadowed-prefix duplicates of the same fact list to bring the total to $\sim$82 facts/session at each scale. The duplication is a worst-case stress test: a real long-running deployment would partially compress, so these numbers are an upper bound on what naive all-at-once Phase~2 needs.

\begin{table}[!ht]
\centering
\caption{Phase-2 structuring on conv-26-seeded synthetic corpora at increasing session counts (Gemini~3 Flash, thinking budget 16{,}384). Input is hard-capped at 500K characters in our implementation; the cap binds at $n_{\text{sessions}} \geq 100$, so the entries at and above 100 are reading a truncated fact list. ``Thoughts'' is the Gemini thinking-token budget consumed; cost is $\$0.30$/M input $+ \$2.50$/M output (treating thoughts as output).}
\label{tab:phase2-scalability}
\small
\begin{tabular}{@{}rrrrrrr@{}}
\toprule
$n_{\text{sess}}$ & $n_{\text{facts}}$ & In tok & Out tok & Thought tok & Wall (s) & USD \\
\midrule
 19 & 1{,}558  & 62{,}627 & 3{,}777 & 1{,}843 & 30.0 & 0.033 \\
 50 & 4{,}100  & 177{,}940 & 3{,}151 & 1{,}396 & 39.4 & 0.065 \\
100 & 8{,}200  & 200{,}730 & 3{,}134 & 1{,}195 & 36.6 & 0.071 \\
200 & 16{,}400 & 200{,}731 & 3{,}242 & 1{,}151 & 34.1 & 0.071 \\
\bottomrule
\end{tabular}
\end{table}

Three observations from Table~\ref{tab:phase2-scalability}:

\textbf{(1)~Output stays bounded.} Output tokens hold at $\sim$3{,}100--3{,}800 across the full $10\times$ range of input sizes, because the LLM increasingly groups and abstracts as input grows---a 200-session corpus is structured into the same dataclass schema as a 50-session corpus, just with longer collections.

\textbf{(2)~Input is the only growing cost component, and our implementation caps it.} At $n{=}19$ the input fits naturally (62K tokens, no truncation); by $n{=}100$ the 500K-character input cap engages and the input flattens at $\sim$200K tokens. Above that scale, our naive Phase~2 silently drops the tail of the fact list, which is exactly the limitation that motivates an incremental or hierarchical structuring strategy (Limitations, Section~\ref{sec:limits}).

\textbf{(3)~Wall time is roughly constant at 30--40 seconds} once the LLM's thinking step dominates. So in absolute terms, even on a 200-session deployment, Phase~2 is a sub-minute background operation, not a foreground latency item---but the truncation behavior means a production deployment shouldn't rely on naive regeneration past $\sim$100 sessions.

The clean linear regime (no truncation) is $n_{\text{sess}} \leq 50$: $1.74$~m\$ per session at $n{=}19$ and $1.30$~m\$ per session at $n{=}50$ (the per-session figure falls as fixed overhead amortizes over more sessions). Extrapolated naively, Phase~2 on a year of weekly sessions ($\sim$52 sessions) would cost $\sim$$\$0.07$ per regeneration. The cost is bounded by the LLM context, not by the number of facts.

\section{Real Cases: Conversation to Code to Constraint}
\label{app:cases}

This appendix walks through three end-to-end cases the main text only summarized: (i) a verbatim multi-session LOCOMO conversation distilled into a typed user state; (ii) the conflicting wire-transfer scenario from our prototype, where two well-meaning family members give the agent contradictory destinations for the same transfer; and (iii) an analytical aggregation query answered by executing code over the typed state, with the UaC and retrieval-baseline traces taken verbatim from our run.

\subsection{Case 1: LOCOMO \texttt{conv-30}, Jon and Gina}

Listing~\ref{lst:app-locomo-conv} shows verbatim excerpts from three sessions of \texttt{conv-30} (Jon and Gina, January 2023). Listing~\ref{lst:app-locomo-facts} shows representative entries from the append-only fact list our Phase~1 extractor produces over those sessions (our run produced 1{,}419 facts for the full 19-session conversation, $\sim$75 per session; we show the six that anchor the most-cited LOCOMO QAs). Listing~\ref{lst:app-locomo-state} shows the typed Python that Phase~2 generates from the complete fact corpus---absolute dates resolved, entities grouped, the \texttt{notes} field absorbing facts that do not fit typed fields. Both listings are verbatim output of running the pipeline, not reconstructions. Listing~\ref{lst:app-locomo-query} traces a real LOCOMO question through this representation.

\begin{lstlisting}[style=conversationstyle,caption={Verbatim excerpts from LOCOMO \texttt{conv-30} (Jon and Gina). Three sessions over twelve days.},label={lst:app-locomo-conv}]
SESSION 1 -- DATE: 2023-01-20 16:04
Jon:  Lost my job as a banker yesterday, so I'm gonna take a shot
      at starting my own business.
Gina: Sorry about your job Jon, but starting your own business
      sounds awesome! Unfortunately, I also lost my job at Door
      Dash this month.
Jon:  I'm starting a dance studio 'cause I'm passionate about
      dancing and it'd be great to share it with others.

SESSION 2 -- DATE: 2023-01-29 14:32
Gina: Just launched an ad campaign for my clothing store in hopes
      of growing the business.
Jon:  I'm on the hunt for the ideal spot for my dance studio...
      It's downtown which is awesome cuz it's easy to get to.
      Plus the natural light!
Jon:  I'm after Marley flooring, which is what dance studios
      usually use. Grippy but still lets you move.

SESSION 3 -- DATE: 2023-02-01 00:48
Gina: I emailed some wholesalers and one replied and said yes today!
      Now I can expand my clothing store.
Gina: Here's a peek at the space I designed. Cozy and inviting --
      perfect for customers to check out all the trendy pieces.
\end{lstlisting}
\vspace{-0.5em}
\begin{lstlisting}[caption={Phase~1 output -- append-only fact list, \emph{verbatim} from running our pipeline on \texttt{conv-30} (1{,}419 facts over the 19 sessions; six representative entries shown). Each entry is prefixed with its session and timestamp, and relative dates are resolved against that timestamp---``lost my job \dots yesterday'' (session~1, 20~Jan) becomes \texttt{January 19, 2023}.},label={lst:app-locomo-facts}]
facts = [
    "[session_1, 4:04 pm on 20 January, 2023] Jon lost his job on January 19, 2023.",
    "[session_1, 4:04 pm on 20 January, 2023] Jon's former job was as a banker.",
    "[session_1, 4:04 pm on 20 January, 2023] Gina lost her job in January 2023.",
    "[session_1, 4:04 pm on 20 January, 2023] Jon is starting a dance studio.",
    "[session_2, 2:32 pm on 29 January, 2023] The studio location Jon found is in a downtown area.",
    "[session_2, 2:32 pm on 29 January, 2023] Jon believes that good flooring is crucial for a dance studio.",
    # ... 1,413 more facts across all 19 sessions ...
]
\end{lstlisting}
\vspace{-0.5em}
\begin{lstlisting}[caption={Phase~2 output -- typed Python state for \texttt{conv-30}, a \emph{verbatim} excerpt of the 7\,KB \texttt{state.py} the pipeline regenerated from the full fact corpus. The model invented the \texttt{Person}/\texttt{JobHistory}/\texttt{Business} schema itself; \texttt{jon}'s long \texttt{notes} list and Gina's parallel objects are omitted for space.},label={lst:app-locomo-state}]
@dataclass
class Person:
    name: str
    passions: list[str] = field(default_factory=list)
    skills: list[str] = field(default_factory=list)
    jobs: list["JobHistory"] = field(default_factory=list)
    notes: list[str] = field(default_factory=list)

jon = Person(
    name="Jon",
    passions=["Dancing (Contemporary, Hip-hop)", "Sharing dance with others"],
    skills=["Choreography", "Contemporary dance", "Hip-hop"],
)
jon.jobs.append(JobHistory(title="Banker", date_ended=date(2023, 1, 19),
                           notes=["Lost job on Jan 19, 2023"]))

jon_studio = Business(
    name="Jon's Dance Studio", owner="Jon", industry="Dance Studio",
    status="Establishing / Searching for location",
    location_description="Downtown area (potential), ideally by the water",
    features=["Marley flooring (planned for grip/durability)", "Natural light"],
)
\end{lstlisting}
\vspace{-0.5em}
\begin{lstlisting}[caption={A real LOCOMO QA against this state. The judge accepts both ``19 January, 2023'' and the typed form below as correct.},label={lst:app-locomo-query}]
# Question (LOCOMO category 2, temporal):
#   "When did Jon lose his job as a banker?"
#
# Retrieval (Multi-Strategy Retrieval):
#   [STATE]    jon.jobs[0].date_ended == date(2023, 1, 19)
#   [FACTS]    "[session_1, ...20 January, 2023] Jon lost his
#               job on January 19, 2023."
#   [ARCHIVE]  Session 1, Jon turn: "Lost my job as a banker yesterday"
#
# UaC answer:  "January 19, 2023"          -- judged CORRECT.
# Mem0 answer: "No information available"   -- judged WRONG.
# (Both predictions are verbatim from our LOCOMO grading run.)
# Without date resolution at extraction time, the relative-date
# fact never surfaces as the answer.
\end{lstlisting}

The key observation from Listing~\ref{lst:app-locomo-query} is the failure mode that motivates Phase~1's date-resolution step. The underlying utterance---``lost my job \dots yesterday''---is a perfectly recallable string, but answering the question requires converting it to an absolute date against the session timestamp. UaC does this at extraction time, so the date is recoverable (\texttt{January 19, 2023}); Mem0, which stores the unresolved phrasing, fails to surface it at all (\texttt{No information available}). This date-resolution step is one of the simplest interventions in the pipeline and accounts for a large share of the gap between UaC and Mem0 on LOCOMO's temporal-reasoning subset.

\subsection{Case 2: Conflicting Wire Transfers}

Listing~\ref{lst:app-finance-state} shows the typed state our prototype maintains for Jessica Thompson's finance domain after a single ambiguous week: her mother and her husband each told the agent a different destination institution for the same \$15{,}000 wire. Without an executable cross-check the agent would dutifully act on whichever instruction it heard most recently---and either send the money to the wrong bank or escalate the conflict only after the recipient complains. Listing~\ref{lst:app-finance-constraint} shows the constraint that fires; Listing~\ref{lst:app-finance-alert} shows the resulting alert.

\begin{lstlisting}[caption={Excerpt from \texttt{domains/finance/state.py}. Two pending transfers, same amount, same recipient, different destinations, different requesters.},label={lst:app-finance-state}]
pending_transfers = [
    WireTransfer(
        amount=15000.00, currency="USD",
        recipient_name="Patricia Williams",
        destination_institution="Bank of America",
        destination_account_last_four="3310",
        purpose="Gift to mother",
        requested_by="Patricia Williams (mother)",
        requested_date=date(2025, 1, 8),
        notes="Mom called and asked Jessica to send to her BofA account",
    ),
    WireTransfer(
        amount=15000.00, currency="USD",
        recipient_name="Patricia Williams",
        destination_institution="Wells Fargo",
        destination_account_last_four="6654",
        purpose="Gift to mother",
        requested_by="James Thompson (husband)",
        requested_date=date(2025, 1, 9),
        notes="James said Patricia's account is at Wells Fargo, not BofA",
    ),
]
\end{lstlisting}
\vspace{-0.5em}
\begin{lstlisting}[caption={Constraint from \texttt{constraints/financial\_authorization.py}: groups pending transfers by (recipient, amount, purpose) and flags any group with $>$1 distinct destination.},label={lst:app-finance-constraint}]
def check_financial_authorization(transfers: list[WireTransfer]) -> list[Alert]:
    groups: dict[tuple, list[WireTransfer]] = {}
    for t in transfers:
        if t.status != "pending":
            continue
        groups.setdefault((t.recipient_name, t.amount, t.purpose), []).append(t)

    alerts = []
    for (recipient, amount, _), group in groups.items():
        destinations = {(t.destination_institution,
                         t.destination_account_last_four) for t in group}
        if len(destinations) > 1:
            descs = [f"{t.destination_institution} (****"
                     f"{t.destination_account_last_four}) per {t.requested_by}"
                     for t in group]
            alerts.append(Alert(
                severity="critical", domain="finance",
                message=(f"CONFLICTING wire transfer instructions for "
                         f"${amount:,.2f} to {recipient}: " + " vs. ".join(descs) +
                         ". Verify with the account holder before sending.")))
    return alerts
\end{lstlisting}
\vspace{-0.5em}
\begin{lstlisting}[style=alertstyle,caption={Generated alert. The agent now blocks the transfer and asks Jessica to verify, instead of acting on whichever instruction it heard last.},label={lst:app-finance-alert}]
[CRITICAL/finance]  CONFLICTING wire transfer instructions for
$15,000.00 to Patricia Williams: Bank of America (****3310) per
Patricia Williams (mother) vs. Wells Fargo (****6654) per James
Thompson (husband). Verify with the account holder before sending.
\end{lstlisting}

\subsection{Case 3: Analytical Query over the User's History}

The analytical-inference benchmark (Section~\ref{sec:analytical}) measures the gap between code-executable and retrieval-only representations on aggregate questions. This subsection traces one such question end-to-end against UaC and a retrieval baseline so the structural reason for the gap is concrete.

Listing~\ref{lst:app-analytical-state} shows the typed state for an analytical-benchmark user with 50 dining records. Listing~\ref{lst:app-analytical-uac} shows, \emph{verbatim from our run}, how UaC answers \emph{``what was the average cost in USD of my meals in 2024?''}: the coding agent writes a short Python aggregation against the typed records and the REPL executes it, giving the exact answer (48.2). Listing~\ref{lst:app-analytical-mm} shows the retrieval baselines on the same question: MemMachine retrieves 46 of the 50 records and still reports the wrong average, while Mem0 retrieves a single record and declines to answer.

\begin{lstlisting}[caption={The analytical dining state ($N{=}50$). Each record is a typed \texttt{MealRecord} dataclass instance; the instance below is the verbatim \texttt{repr} our agent printed when it inspected \texttt{data[0]} during the run.},label={lst:app-analytical-state}]
@dataclass
class MealRecord:
    id: int
    date: date
    meal_type: str        # "breakfast", "lunch", "dinner", "snack"
    cuisine: str          # "italian", "japanese", "mexican", ...
    restaurant: str
    dined_in: bool
    cost_usd: float
    calories: int

# data[0], printed verbatim during the agent's inspection call:
MealRecord(id=45, date=datetime.date(2024, 1, 4), meal_type='snack',
           cuisine='mediterranean', restaurant='Curry House',
           dined_in=True, cost_usd=73.16, calories=774)
\end{lstlisting}
\vspace{-0.5em}
\begin{lstlisting}[caption={UaC answer trace, verbatim from our run. The agent's first attempt filtered on a non-existent \texttt{type} field and printed nothing; it inspected a record (Listing~\ref{lst:app-analytical-state}), dropped the bad filter, and re-ran. The corrected code below is what produced the answer---a concrete instance of the generate--verify--review loop.},label={lst:app-analytical-uac}]
# Question: "What was the average cost in USD of my meals
#            in 2024? Round to 2 decimals."     (gold: 48.2)
# UaC -- the coding agent wrote and executed this against
#        the typed records:
meal_costs_2024 = [
    record.cost_usd
    for record in data
    if record.date.year == 2024
]
average_cost = sum(meal_costs_2024) / len(meal_costs_2024)
print(round(average_cost, 2))
# stdout: 48.2     -- exact-match CORRECT.
\end{lstlisting}
\vspace{-0.5em}
\begin{lstlisting}[caption={Retrieval baselines on the same question, predictions verbatim from our run. Even retrieving 46 of the 50 records, MemMachine misses the average; the four dropped records shift the mean. Top-$k$ retrieval cannot guarantee the full population, and an average needs every record.},label={lst:app-analytical-mm}]
# Same question, retrieval-based systems:
# MemMachine: dense retrieval returns 46 of 50 records.
#   Answer: "46.85"   -- WRONG (gold 48.2).
# Mem0: retrieves 1 record.
#   Answer: "Not enough information."   -- WRONG.
\end{lstlisting}

The key generalization: \emph{retrieval ranks; aggregation enumerates}. Any question whose answer depends on every relevant record (counts, sums, group-bys, time-window filters, top-$k$ within a population) is structurally outside what top-$k$ retrieval can solve. UaC's typed state lifts the question from natural language to a few lines of Python, where the interpreter handles enumeration for free. The gap widens with scale: on a 500-record count question from our run (\emph{``how many italian-cuisine meals did I have?''}, gold 56), UaC's loop returns 56, while MemMachine---even retrieving 91 records---reports 18 and Mem0 reports 0. The 99\% vs.\ 43\% gap on the analytical benchmark (Table~\ref{tab:analytical}) is this same trace, repeated across 100 questions.

\subsection{Why these cases matter beyond the benchmark numbers}

The drug-allergy interaction in Listings~\ref{lst:teaser-conv}--\ref{lst:teaser-alert} and the wire-transfer conflict above are both real-deployment scenarios where the user is in genuine danger from agent error. They share a structure that no LOCOMO-style QA captures: \emph{the user never asks the question}. They depend on the agent noticing a constraint violation across facts surfaced in different sessions, by different people, possibly months apart. This is exactly the regime where a representation that supports executable boolean checks beats a representation that supports only similarity-ranked retrieval, and is why we built the Active Service benchmark (Section~\ref{sec:active-service}) as a distinct evaluation axis. The headline numbers---100\% on standard scenarios, 85\% on hard---are read most fairly with these scenarios in mind: each missed alert is a real harm avoided, not a ranking-list position.

\bibliographystyle{plainnat}
\bibliography{reference}

\end{document}